\documentclass[lettersize,journal,table]{IEEEtran}
\usepackage{amsmath,amsfonts}
\usepackage{algorithmic}
\usepackage{algorithm}
\usepackage{array}
\usepackage[caption=false,font=normalsize,labelfont=sf,textfont=sf]{subfig}
\usepackage{caption}
\usepackage{textcomp}
\usepackage{stfloats}
\usepackage{verbatim}
\usepackage{graphicx}
\usepackage{cite}
\usepackage{multirow}
\usepackage{multicol}
\usepackage{pgfplots}
\usepackage{booktabs}
\usepackage{tikz}
\usetikzlibrary{positioning}

\usepackage{url}
\usepackage{xcolor}
\usepackage[colorlinks=true, citecolor=blue, linkcolor=blue, urlcolor=blue]{hyperref}

\begin{document}

\title{Few-shot Class-incremental Fault Diagnosis by Preserving Class-Agnostic Knowledge with Dual-Granularity Representations}

\author{Zhendong Yang, Jie Wang$^{\star}$, Liansong Zong, Xiaorong Liu, Quan Qian, Shiqian Chen

\thanks{$^{\star}$Corresponding author: Jie Wang.}

\thanks{Zhendong Yang is with the School of Computer and Software Engineering, Xihua University.}
\thanks{Jie Wang is with the School of Computing and Artificial Intelligence, Southwest Jiaotong University, Chengdu, China. Email: jackwang@swjtu.edu.cn.}
\thanks{Liansong Zong is with the School of Computing and Artificial Intelligence, Southwest Jiaotong University, Chengdu, China, and also with the School of Computer and Software Engineering, Xihua University, Chengdu, China.}
\thanks{Xiaorong Liu is with the School of Computer Science and Artificial Intelligence, Southwest Petroleum University, Chengdu, China.}
\thanks{Quan Qian is with the School of Automation Engineering, University of Electronic Science and Technology of China, Chengdu, China.}
\thanks{Shiqian Chen is with the State Key Laboratory of Rail Transit Vehicle System, Chengdu, China.}
\thanks{This work is supported by the Fundamental Research Funds for the Central Universities (Grant No. 2682025CX105) and the National Natural Science Foundation of China (Grant No. 62406259).}
}

\maketitle

\begin{abstract}
Few-Shot Class-Incremental Fault Diagnosis (FSC-FD), which aims to continuously learn from new fault classes with only a few samples without forgetting old ones, is critical for real-world industrial systems. However, this challenging task severely amplifies the issues of catastrophic forgetting of old knowledge and overfitting on scarce new data. To address these challenges, this paper proposes a novel framework built upon Dual-Granularity Representations, termed the Dual-Granularity Guidance Network (DGGN). Our DGGN explicitly decouples feature learning into two parallel streams: 1) a fine-grained representation stream, which utilizes a novel Multi-Order Interaction Aggregation module to capture discriminative, class-specific features from the limited new samples. 2) a coarse-grained representation stream, designed to model and preserve general, class-agnostic knowledge shared across all fault types. These two representations are dynamically fused by a multi-semantic cross-attention mechanism, where the stable coarse-grained knowledge guides the learning of fine-grained features, preventing overfitting and alleviating feature conflicts. To further mitigate catastrophic forgetting, we design a Boundary-Aware Exemplar Prioritization strategy. Moreover, a decoupled Balanced Random Forest classifier is employed to counter the decision boundary bias caused by data imbalance. Extensive experiments on the TEP benchmark and a real-world MFF dataset demonstrate that our proposed DGGN achieves superior diagnostic performance and stability compared to state-of-the-art FSC-FD approaches. Our code is publicly available at \url{https://github.com/MentaY/DGGN}.
\end{abstract}


\section{Introduction}

\IEEEPARstart{I}{ntelligent} fault diagnosis is pivotal for ensuring the operational reliability and safety of modern industrial systems by identifying equipment failures from sensor data. While data-driven methods have shown remarkable
\begin{figure}[!htbp]
    \includegraphics[width=\columnwidth]{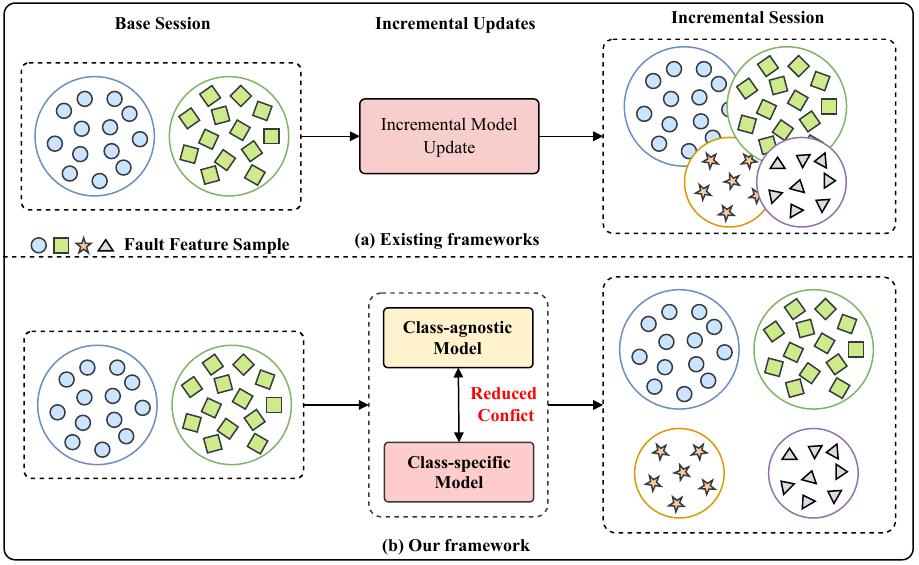}
      \centering
      \caption{(a) Existing methods often suffer from feature conflicts when integrating new fault classes. (b) Our framework introduces a class-agnostic and a class-specific model to collaboratively reduce feature conflicts and enhance fault separability.}
    \label{fig1}
\end{figure}
\noindent success \cite{an2023gaussian, gao2022hierarchical, li2023order, wang2024hard, wang2024generalized}, their efficacy often relies on a static, data-rich training environment. This assumption is frequently violated in practice, as industrial systems operate predominantly under normal conditions, rendering fault samples inherently scarce. More critically, fault patterns are not static and evolve dynamically due to factors such as changing operating conditions and component aging. This necessitates diagnostic models that can incrementally learn new fault types without requiring complete retraining.

These practical constraints motivate the study of Few-Shot Class-Incremental Fault Diagnosis (FSC-FD). This challenging task amplifies two core problems for diagnostic models: the catastrophic forgetting of previously acquired knowledge and the severe overfitting on the scarce samples of new fault classes \cite{mccloskey1989catastrophic}. Developing a methodology that concurrently addresses this dual challenge is crucial for the deployment of truly adaptive intelligent diagnosis systems in real-world settings.

Recent works in Class-Incremental Fault Diagnosis (CIL-FD) have explored techniques such as contrastive learning \cite{li2023incrementally, peng2023sclifd}, lifelong learning frameworks \cite{chen2022lifelong, bojian2023continual}, and memory replay strategies \cite{zheng2022bearing}. Despite this progress, a fundamental bottleneck persists, particularly in the few-shot context. Existing methods typically attempt to learn and balance knowledge within a single, monolithic feature space. This approach leads to representation entanglement as new classes are added, where discriminative features of new classes interfere with the established representations of old ones. As illustrated in \figurename~\ref{fig1}, this entanglement causes feature confusion and biases the model's decision boundaries toward data-rich majority classes, hindering effective generalization for new, minority fault types.

To overcome this limitation, this paper proposes a novel framework that decouples feature learning through what we term Dual-Granularity Representations. Our proposed Dual-Granularity Guidance Network (DGGN) explicitly learns two parallel representations within a unified architecture. It comprises a fine-grained, class-specific stream designed to capture subtle, discriminative features from limited new samples using a novel Multi-Order Interaction Aggregation (MOIA) module. In parallel, a coarse-grained, class-agnostic stream continuously models the general, foundational knowledge shared across all fault types, preserving a stable representation of what constitutes a fault. A key innovation is our Multi-Semantic Cross-Attention mechanism, which enables the stable coarse-grained knowledge to guide and regularize the learning of the fine-grained representations, thereby preventing overfitting and alleviating feature conflicts. Furthermore, a Boundary-Aware Exemplar Prioritization (BAEP) strategy is developed to optimize the replay mechanism against forgetting. Finally, a decoupled Balanced Random Forest (BRF) classifier is trained on the learned features to mitigate classification bias arising from data imbalance. The main contributions of this paper are summarized as follows:

\begin{itemize}
    \item We proposes a novel framework built upon Dual-Granularity Representations, which models class-specific features and class-agnostic semantics in parallel. A multi-semantic cross-attention mechanism is introduced to dynamically fuse features from both semantic domains and transfer the unified representation back to the class-specific model, effectively mitigating feature conflicts and enhancing feature representation under few-shot scenarios.
    \item We design a Multi-Order Interaction Aggregation module to capture multi-order contextual interactions from limited fault signals. Additionally, we introduce a Boundary-Aware Exemplar Prioritization strategy to mitigate forgetting and adopt a Balanced Random Forest classifier to address classification bias caused by class imbalance.
    \item We conduct extensive experiments on the TEP benchmark and real-world MFF datasets, and the results demonstrate that the proposed DGGN framework achieves significant improvements over baseline models in both diagnostic accuracy and stability.
\end{itemize}

\section{Related Work}

\subsection{Class incremental learning}
Class-Incremental Learning (CIL), a crucial subfield of continual learning, enables models to sequentially learn new classes while preserving performance on previously seen ones \cite{zhou2024class}. Prevailing CIL methodologies can be broadly classified into three categories. \textit{1) Replay-based methods} \cite{belouadah2019il2m, hou2019learning, rebuffi2017icarl, zhu2021prototype} mitigate catastrophic forgetting by storing a small memory buffer of old class exemplars and replaying them during incremental training steps. \textit{2) Regularization-based methods} \cite{chen2022multi, liu2018rotate, yang2021learning} impose constraints to penalize significant updates to parameters deemed critical for past tasks, thereby preserving learned knowledge. \textit{3) Architecture-based methods} \cite{yang2021incremental, zhu2021class} dynamically adapt the model's architecture, such as by isolating or expanding network components, to accommodate new classes without interfering with old knowledge.
However, the efficacy of these traditional CIL methods is severely challenged under the more realistic and demanding Few-Shot Class-Incremental Learning (FSCIL) setting. In this scenario, the model must learn from a mere handful of samples for each new class. This data scarcity not only exacerbates catastrophic forgetting but also introduces a high risk of overfitting to the sparse new data, leading to poor generalization. While some advanced techniques have been proposed, such as learning in orthogonal subspaces \cite{chaudhry2020continual}, using Hadamard products for efficient weight composition \cite{wen2020batchensemble}, or aligning gradients between new and old tasks \cite{xiang2023tkil}, they still often learn a monolithic feature representation.
A primary limitation of these existing approaches is their failure to explicitly decouple general, class-agnostic knowledge from discriminative, class-specific features. They tend to entangle these two types of semantics, making it difficult to learn robust new concepts from scarce data without disrupting the stable, foundational knowledge learned from previous tasks. We argue that by preserving and leveraging this stable class-agnostic knowledge, it can serve as a powerful guide for learning new class-specific representations. This motivates our work to develop a framework built on dual-granularity representations to address this critical gap.

\subsection{Fault Diagnosis}
With the advancement of intelligent manufacturing, data-driven fault diagnosis methods \cite{wang2025diagllm} attract widespread attention in both academia and industry. Deep learning approaches in particular, such as CNN \cite{xia2017fault}, RNN \cite{liu2018fault}, and LSTM \cite{mohammad2023one}, demonstrate remarkable capabilities in modeling the complex nonlinear relationships between condition monitoring data and equipment health status, thereby achieving accurate fault diagnosis. However, most of these methods operate on the assumption that all fault categories are known and fixed during the training phase. This makes them unable to handle novel fault types that continuously emerge in real-world settings and causes them to suffer from ``catastrophic forgetting.'' Consequently, conventional deep learning models are ill-suited for dynamically evolving industrial environments.

To address these challenges, researchers have proposed methods based on class-incremental learning. For instance, Kirkpatrick et al. \cite{kirkpatrick2017overcoming} introduce the Elastic Weight Consolidation module, which estimates parameter importance to restrict model updates. Aljundi et al. \cite{aljundi2018memory} propose the Memory Aware Synapses approach to evaluate parameter sensitivity based on function gradients. Li et al. \cite{li2017learning} utilize knowledge distillation to guide the student network to preserve information learned by a teacher model. On the other hand, Rebuffi et al. \cite{rebuffi2017icarl} present a prioritization-based sample selection strategy for memory replay. Shin et al. \cite{shin2017continual} introduce Generative Adversarial Networks (GANs) to simulate and retain old class data by jointly training generator and discriminator modules. Chen et al. \cite{chen2022lifelong} propose a dual-branch aggregation network to balance the stability and plasticity of deep neural networks, thereby improving knowledge retention for previously seen classes.

Although significant progress has been made, existing methods still face limitations when confronted with real-world industrial challenges such as \textit{limited samples}, \textit{class imbalance} and \textit{semantic conflict}. Particularly at the representation learning level, they lack the ability to perform unified semantic modeling for both old and new tasks. In this paper, our proposed dual-granularity representation framework integrates class-specific and class-agnostic representations. It introduces a multi-semantic cross-attention mechanism for dynamic fusion and leverages the generalizable knowledge from the class-agnostic domain to guide the learning of the class-specific model. This approach enhances the incremental diagnosis capability, especially under class imbalance conditions.

\section{Method}
\subsection{Preliminaries}

\textbf{Problem Definition.} Class-Incremental Learning for Fault Diagnosis aims to develop intelligent diagnostic models capable of continuously learning new fault categories from non-stationary data streams, while simultaneously retaining the knowledge of previously seen classes. This learning paradigm typically involves two main phases: (1) Base Session Training, and (2) Incremental Session Training. During the base session, the model learns from sufficient and balanced training data to acquire initial knowledge of fault classes. In the incremental sessions, new fault categories are introduced with limited data samples, which often exhibit class imbalance and long-tailed distributions. Formally, the CIL-FD task is defined as a sequence of $n$ learning sessions:
\begin{equation}\mathcal{T} = \left[(C^1, D^1), (C^2, D^2), \ldots, (C^n, D^n)\right],\end{equation}
where each session $t$ consists of a class set $C^t = \left\{ c_1^t, c_2^t, \ldots, c_{n_t}^t \right\}$ and its corresponding training data $D^t$. In the initial session, the training dataset is denoted as $D^1 = \left\{ (x_1, y_1), (x_2, y_2), \ldots, (x_i, y_i), y_i \in C^1 \right\}$, where $x_i$ represents a fault signal sample and $y_i$ is its corresponding class label. In each subsequent incremental session $t> 1$, the new dataset is represented as $D^t = \left\{ (x_i, y_i)_{i=1}^N, y_i \in C^t \right\}$ denotes the set of new classes introduced at session $t$. To ensure a clear separation of learning tasks, the class sets across different sessions are disjoint, i.e., $\forall t_1 \neq t_2, C^{t_1} \cap C^{t_2} = \emptyset$. During the evaluation phase, the model is expected to recognize samples from all the learned classes. Therefore, the final diagnostic model should be able to correctly classify inputs drawn from the cumulative class set: 
\begin{equation}C^{(0:n)} = \left\{ C^1 \cup C^2 \cup \cdots \cup C^n \right\},\end{equation}
and demonstrate consistent performance across all fault categories.

\textbf{Feature Diversity.} In single-task fault diagnosis classification, the model only needs to capture the minimal set of discriminative features required to complete the task. For example, in a binary classification task involving only ``severe outer ring wear of the bearing '' and ``normal motor operation,'' it is sufficient for the model to detect whether the energy amplitude in a specific high-frequency band of the vibration signal significantly increases. This enables the model to distinguish the two conditions with high accuracy while minimizing classification loss. These features, which are highly discriminative for the current classification objective, are often only effective within the current class distribution and struggle to generalize to newly introduced classes. We refer to them as class-specific features. However, beyond these task-specific features, the data also contain a large number of latent features that are not explicitly used for classification. These features may not directly correspond to specific fault categories but can reflect the underlying operating conditions of the equipment, environmental noise, or the inherent physical properties of the signal. We define these as class-agnostic features, and this paper explores how jointly modeling these two feature types can enhance both the performance and generalization capability of class-incremental fault diagnosis.

\begin{figure*}[!htbp]
    \includegraphics[width=1\textwidth]{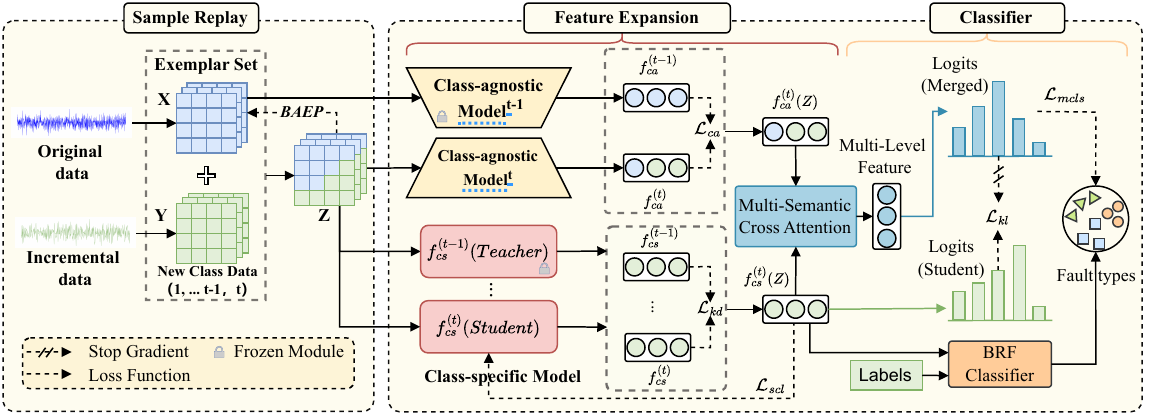}
      \centering
      \caption{Architecture of the proposed Dual-Granularity Guidance Network. The framework comprises three key components: \textit{sample replay}, \textit{feature expansion}, and \textit{classifier}.  The Boundary-Aware Exemplar Prioritization strategy selects representative samples to mitigate catastrophic forgetting. In the core module, class-agnostic knowledge is preserved to guide the learning of class-specific representations, thereby preventing feature entanglement. Finally, a decoupled Balanced Random Forest classifier is used to address data imbalance during classification.}
    \label{fig2}
\end{figure*}

\subsection{Approach Overview}       

The overall architecture of our approach is illustrated in \figurename~\ref{fig2}. Specifically, our method consists of four core components: class-specific representation learning, class-agnostic guided feature expansion, dual-domain feature fusion and alignment, and decoupled classification.

We begin by extracting task-relevant features from fault signals using a dual-branch feature extractor. To enhance the discriminability of learned features, we design a Multi-Order Interaction Aggregation module that captures both local and global information through hierarchical interactions. To further mitigate the challenges of catastrophic forgetting and class imbalance, we incorporate knowledge distillation within the class-specific branch via a teacher-student paradigm, and introduce a sample replay mechanism based on the Boundary-Aware Exemplar Prioritization strategy.

Meanwhile, we deploy a class-agnostic branch to continuously model the shared semantic information across tasks. A Multi-Semantic Cross Attention module is proposed to align and integrate class-agnostic and class-specific features, enabling the fused representations to guide the optimization of the class-specific learner. Finally, to decouple the complexity of feature learning from the classification decision, we employ a separately trained BRF classifier to make the final predictions.

\subsection{Class-Specific Representation Learning} 

To effectively capture discriminative patterns of newly introduced fault classes under limited data conditions, we construct a class-specific feature extraction branch as the primary semantic encoding pathway for fault information. This branch is continuously updated during each incremental session to extract discriminative features relevant to the currently introduced classes. To further enhance its representational capacity, we design a MOIA module and incorporate feature-level knowledge distillation to alleviate catastrophic forgetting. 
\begin{figure}[!htbp]
    \includegraphics[width=\columnwidth]{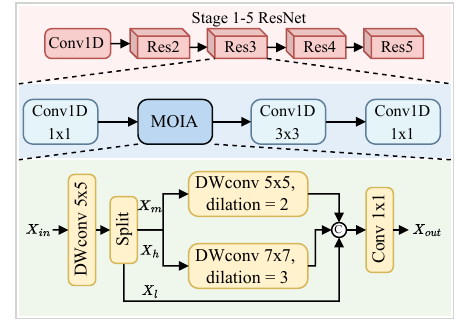}
      \centering
      \caption{Class-Specific Model Architecture with Multi-Order Interaction Aggregation Module.}
    \label{fig3}
\end{figure}

\noindent This section details the update procedure of the class-specific model within our proposed framework.

\subsubsection{Multi-Order Interaction Aggregation Module.} 

As illustrated in \figurename~\ref{fig3}, we adopt 1D ResNet-18 as the backbone of the class-specific model and incorporate a Multi-Order Interaction Aggregation module into the residual blocks. This module encodes multi-order features within the class-specific branch via depthwise separable convolutions (DWConv). Different from prior works \cite{lu2022unbalanced, zhang2024multiscale} that simply combine DWConv with attention to capture contextual representations, we utilize three parallel DWConv layers with dilation rates $d \in \{1, 2, 3\}$ to extract low-, mid-, and high-level contextual features. Specifically, given an input feature $X_{in} \in \mathbb{R}^{C \times L}$, where $C$ denotes the number of channels and $L$ the sequence length, we first apply a $\text{DWConv}_{5 \times 5}$ with dilation $d=1$ to obtain the initial encoding. The output is then split along the channel dimension into three parts: low-level features $X_l \in \mathbb{R}^{C_l \times L}$,  mid-level features $X_m \in \mathbb{R}^{C_m \times L}$, and high-level features $X_h \in \mathbb{R}^{C_h \times L}$. The mid- and high-level features are further fed into DWConv layers with dilation rates 2 and 3, respectively, yielding the following:
\begin{equation}\tilde{X}_m = DW_{5 \times 5, d=2}(X_m),\end{equation}
\begin{equation}\tilde{X}_h = DW_{7 \times 7, d=3}(X_h),\end{equation}

\noindent where $\tilde{X}_m$ and $\tilde{X}_h$ extract mid- and high-range contextual representations, respectively, which are crucial for capturing cross-scale dependencies in complex fault scenarios. Subsequently, the three features $X_l$, $\tilde{X}_m$ and $\tilde{X}_h$ are concatenated along the channel dimension and compressed through a $1 \times 1$ convolution to produce the final multi-order contextual representation $x_out$, as formulated in (5):
\begin{equation}X_{out} = Conv_{1 \times 1} \left( Concat(X_l, \tilde{X}_m, \tilde{X}_h) \right).\end{equation}
The resulting feature $X_{out} \in \mathbb{R}^{C \times L}$ effectively integrates both local details and global semantics from fault signals, significantly enhancing the representational capacity of the class-specific model under limited data conditions.

\subsubsection{Supervised Contrastive Knowledge Distillation.}

To ensure that the features extracted by the current encoder $f_{cs}^{(t)}$ exhibit high intra-class compactness and inter-class separability, we adopt a supervised contrastive learning strategy \cite{khosla2020supervised}. As shown in \figurename~\ref{fig4}, the core idea is to encourage embeddings of samples from the same class (i.e., positive samples) to cluster together in the embedding space, while simultaneously pushing apart those from different classes (i.e., negative samples). Given a mini-batch of labeled samples $(x_i, y_i) \in D_t$, and the corresponding embeddings $z_i = f_{cs}(x_i)$, the supervised contrastive loss is defined as:
\begin{figure}[!htbp]
    \includegraphics[width=\columnwidth]{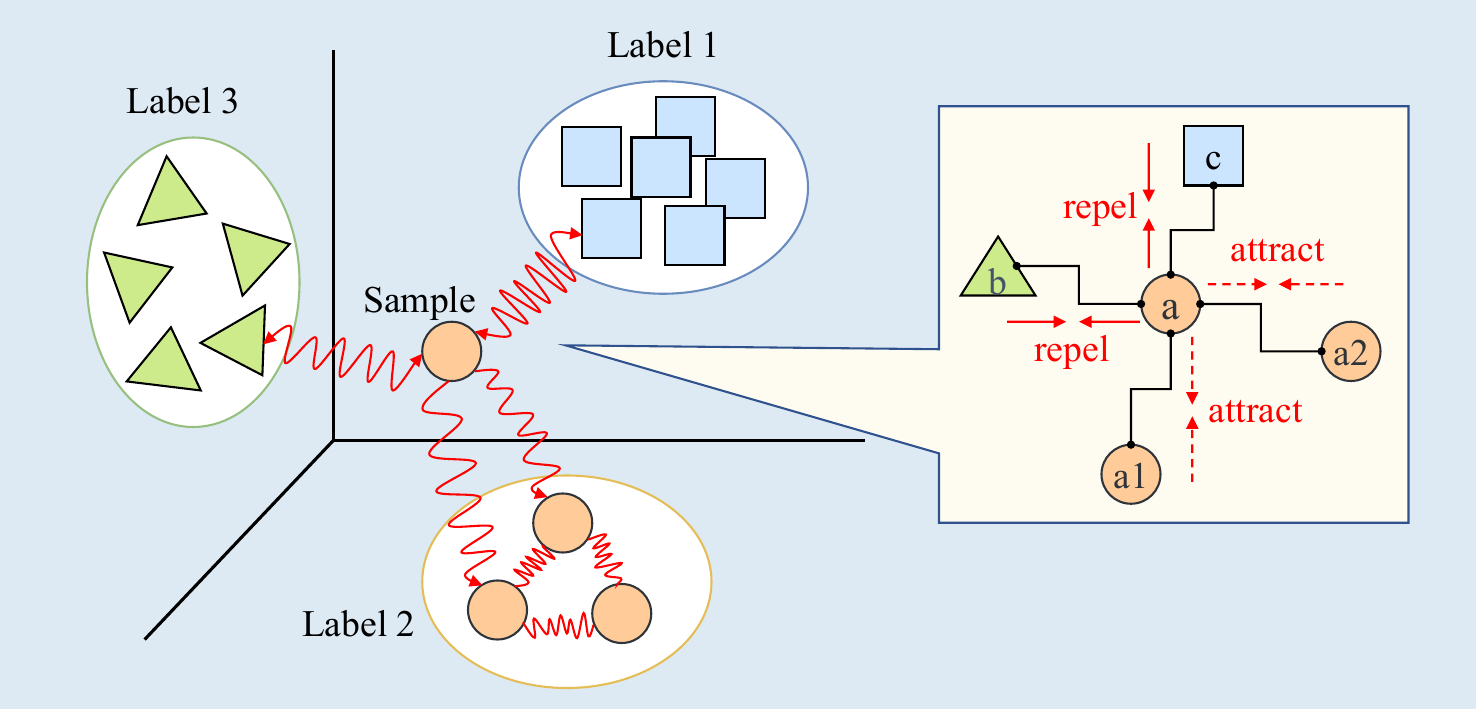}
      \centering
      \caption{Interactions between instances and labels in contrastive learning.}
    \label{fig4}
\end{figure}
\begin{equation}\mathcal{L}_{\text{scl}} = \sum_{i \in I} \frac{-1}{|P(i)|} \sum_{p \in P(i)} \log \frac{\exp\left( (z_i \cdot z_p)/\tau \right)}{\sum_{a \in A(i)} \exp\left( (z_i \cdot z_a)/\tau \right)},\end{equation}
where $\tau \in \mathbb{R}^+$ denotes a temperature parameter, and the dot product $\cdot$ represents the inner product. $\text{I}$ is the index set of all samples, and $A(i) \equiv I \setminus \{i\}$ represents the set of all anchor samples excluding $x_{i}$. The set $P(i) \equiv \{ p \in A(i) : y_p = y_i \}$ includes all positive samples that share the same label as $x_{i}$, and $|P(i)|$ denotes the number of such samples.

To further mitigate the effects of catastrophic forgetting, we introduce a feature-level knowledge distillation method. Unlike traditional approaches that distill softmax-based class distributions \cite{rebuffi2017icarl}, our method aligns the feature representations between the student and teacher models in the embedding space. Specifically, we freeze the feature encoder $f_{cs}^{(t-1)}$ from the previous session as the teacher model and optimize the current session's encoder $f_{cs}^{(t)}$ as the student. The teacher provides guidance based on its prior distribution over old classes, and the distillation is performed via a contrastive similarity loss defined as:
\begin{equation}
Q(z_i; z_a) = \frac{\exp\left( (z_i \cdot z_a)/\tau \right)}{\sum_{k \in A(i)} \exp\left( (z_i \cdot z_k)/\tau \right)},
\end{equation}

\begin{equation}
\begin{split}
\mathcal{L}_{\text{kd}}(f_{\text{cs}}^{t-1}, f_{\text{cs}}^t) &= -\frac{1}{2N} \sum_{i \in I} \sum_{a \in A(i)} Q(z_i; z_a)^T \log \left( Q(z_i; z_a)^S \right) \\
&= -\frac{1}{2N} \sum_{i \in I} \sum_{a \in A(i)} \Bigg[ Q\big(f_{\text{cs}}^{t-1}(x_i); f_{\text{cs}}^{t-1}(x_a)\big) \\
&\quad\quad\quad\log \left( Q\big(f_{\text{cs}}^t(x_i); f_{\text{cs}}^t(x_a)\big) \right) \Bigg]
\end{split}
,\end{equation}
where $Q(z_i; z_a)$ measures the similarity between the anchor feature $z_{i}$ and the auxiliary feature $z_{a}$. 
The superscripts $S$ and $T$ represent the student and teacher models, respectively. $z_{i}$ and $z_{a}$ correspond to features extracted from samples $x_{i}$ and $x_{a}$ and are normalized before similarity is computed. The term $\text{2N}$ denotes the number of sample pairs in a mini-batch, accounting for bidirectional similarity matching.

\subsection{Class-Agnostic Guided Feature Expansion} 
The primary goal of the class-specific branch is to discriminate among newly introduced fault classes in each task. However, when the sample size of these new categories is limited, the model tends to overfit the most discriminative patterns for the new classes. This inclination may result in learned features that lack generalization and are entangled with those of previously seen classes in the feature space. To mitigate this issue, we introduce a parallel class-agnostic feature branch that complements the class-specific encoder by capturing class-independent representations. These general representations serve as shared semantic priors across tasks to guide and regularize the learning of the class-specific model.

Specifically, to continuously acquire task-agnostic features, we design a standalone encoder $f_{ca}$. This encoder leverages the prediction-based self-supervised learning approach from CaSSLe \cite{fini2022self} to extract more general features. Furthermore, we integrate both SimCLR \cite{chen2020simple} and CaSSLe as base learning objectives to obtain a ``semantic anchor'' for each session. As illustrated in the right half of \figurename~\ref{fig2} (Class-Specific Model), we use an independent model to continuously maintain class-agnostic representations. The backbone of $f_{ca}$ remains a 1D ResNet-18, the same as that of $f_{cs}$, but it is trained to learn generalizable representations over all previously seen data in each incremental session.

Given an input sample $x_{i}$, we generate two different augmented views $x_{i}$ and $x_i'$, which are embedded via the current class-agnostic encoder as $z_i = f_{ca}(x_i)$ and $z_i' = f_{ca}(x_i')$. We employ an InfoNCE loss to maximize the mutual information between these embeddings during the base session. The loss is formulated as:
\begin{equation}\begin{split}
\mathcal{L}_{\text{ca}}^{(t=0)} &= -\text{InfoNCE} \, \mathbb{E} \left( \{ f_{\text{ca}}(x_i) \}_{i=1}^N, \{ f_{\text{ca}}(x_i') \}_{i=1}^N \right) \\
&= \frac{1}{N} \sum_{i=1}^N \log \frac{\exp\left( \text{sim}(z_i, z_i') / \tau \right)}{\sum_{\substack{j=1 \\ j \neq i}}^N \exp\left( \text{sim}(z_i, z_j') / \tau \right)}
\end{split},\end{equation}

\begin{equation}\text{sim}(z_i, z_i') = \frac{z_i^\top z_i'}{\| z_i \| \cdot \| z_i' \|},\end{equation}

where $N$ is the batch size, $\text{sim}(z_i, z_i') $ denotes the normalized cosine similarity, and $\tau$ is a temperature parameter. This self-supervised contrastive objective encourages $f_{ca}$ to learn general-purpose representations that are not bound to class-specific labels.

During incremental sessions $(t> 0)$, to maintain semantic consistency in the representation space, we retain the InfoNCE objective while introducing an additional cross-session prediction alignment loss. Unlike the class-specific model $f_{cs}$, which is updated for each session, the class-agnostic encoder $f_{ca}$ is frozen after each session and preserved as a semantic anchor for the following sessions. In each incremental step, we introduce a lightweight predictor $g(\cdot)$ that maps the current class-agnostic embedding to the previous session’s representation space $f_{\text{ca}}^{t-1}(x_i)$. The complete objective for class-agnostic learning becomes:
\begin{equation}\begin{split}
\mathcal{L}_{\text{ca}}^{(t>0)} &= -\text{InfoNCE} \, \mathbb{E} \left( \{ f_{\text{ca}}^t(x_i) \}_{i=1}^N, \{ f_{\text{ca}}^t(x_i') \}_{i=1}^N \right) \\
&\quad -\text{InfoNCE} \, \mathbb{E} \left( \{ g(f_{\text{ca}}^t(x_i)) \}_{i=1}^N, \{ f_{\text{ca}}^{t-1}(x_i) \}_{i=1}^N \right)
\end{split},\end{equation}
where $f_{\text{ca}}^{t-1}(\cdot)$ represents the frozen class-agnostic encoder from the previous session. The core idea of the predictor is to minimize the divergence between current and previous embeddings. If this divergence is minimized, the current features are guaranteed to retain at least as much semantic information as those in the previous session. Therefore, as the learning progresses, the class-agnostic encoder gradually accumulates general representations across all tasks.

\subsection{Guided Dual-Granularity Fusion and Alignment} 

\subsubsection{Multi-Semantic Cross Attention.} 

To alleviate the discrepancy in semantic distribution between the class-specific features learned from the current task and the class-agnostic features inherited from previous tasks, we propose a multi-semantic cross-attention mechanism. This design adaptively controls the feature fusion process and fully exploits the potential of class-agnostic information. Given the class-specific feature map $F_{\text{cs}}^t(\boldsymbol{x})$ and the class-agnostic feature map $F_{\text{ca}}^t(\boldsymbol{x})$, we first normalize both features as follows:
\begin{equation}z_{\text{cs}} = \text{LayerNorm}_1 \bigl( F_{\text{cs}}^t(x) \bigr),\end{equation}
\begin{equation}z_{\text{ca}} = \text{LayerNorm}_2 \Bigl( \text{StopGradient} \bigl( F_{\text{ca}}^t(x) \bigr) \Bigr).\end{equation}

To leverage the useful information from class-agnostic features, we treat $z_{cs}$ as the query vector, which enables it to attend to both $z_{cs}$ and $z_{ca}$. Since $z_{cs}$ and $z_{ca}$ reside in distinct feature spaces, we compute their keys and values separately as:
\begin{equation}Q = W_q \cdot z_{\text{cs}}, \quad K_{\text{cs}} = W_k^{\text{cs}} \cdot z_{\text{cs}}, \quad V_{\text{cs}} = W_v^{\text{cs}} \cdot z_{\text{cs}},\end{equation}
\begin{equation}K_{\text{ca}} = W_k^{\text{ca}} \cdot z_{\text{ca}}, \quad V_{\text{ca}} = W_v^{\text{ca}} \cdot z_{\text{ca}},\end{equation}
where $W_{q}$, $W_{k}$, $W_{v}$ denote learnable linear projections. We then concatenate keys and values from both branches and apply multi-head scaled dot-product attention to obtain the final multi-semantic fusion representation:
\begin{equation}z_m^{(h)} = \text{Softmax} \left( \frac{Q^{(h)} \left[ K_{\text{cs}}^{(h)}, K_{\text{ca}}^{(h)} \right]^T}{d/h} \right) \left[ V_{\text{cs}}^{(h)}, V_{\text{ca}}^{(h)} \right],\end{equation}
where $d$ is the embedding dimension and $h$ is the number of attention heads. This mechanism introduces semantic priors in an adaptive manner, effectively expanding the representation space of current task features. Consequently, the fused representation captures not only current task semantics but also general knowledge across tasks.

Although the fused representation incorporates complementary semantics from both branches, unconstrained fusion may lead to bias away from class-specific learning objectives. To address this, we apply global average pooling (GAP) and use cross-entropy loss to supervise the classification of the fused features:
\begin{equation}\mathcal{L}_{\text{mcls}} = \ell_{\text{CE}}(\text{GAP}(z_m), y),\end{equation}
where $\ell_{\text{CE}}$ is the cross-entropy loss and $y$ denotes the ground-truth label. It is important to note that gradients are not propagated back to the class-agnostic feature extractor, ensuring that the class-agnostic representation remains independent from task-specific discriminative supervision.

\subsubsection{Cross-Domain Knowledge Alignment.}

Through the multi-semantic cross-attention mechanism, the model captures more diverse feature representations. However, as the class-agnostic model continues to be updated, semantic drift may occur. To mitigate this issue, we do not directly rely on the fused features for final prediction. Instead, we transfer the diversified information in the fused features to the feature space of the class-specific task.

Specifically, we feed the fused feature into a classifier to obtain the predicted class probability distribution $p_{m}$ through a fully connected layer. Simultaneously, the class-specific branch $f_{cs}^{\left ( t \right ) }$ also produces its own classification output $p_{cs}$. To enable the fused features to transfer shared semantic knowledge across domains to the class-specific model, we minimize the Kullback-Leibler (KL) divergence between these two distributions. The corresponding loss function is defined as:
\begin{equation}\mathcal{L}_{kl} = \sum_{i=1}^c \text{SG}(p_m(i)) \log \left( \frac{\text{SG}(p_m(i))}{p_{\text{cs}}(i)} \right),\end{equation}
where SG denotes the StopGradient operation, which blocks the gradient flow from the fused features. This mechanism facilitates the transfer of semantic knowledge from the class-agnostic domain to the class-specific domain.
During incremental sessions, we jointly minimize the supervised contrastive loss $\mathcal{L}_{scl}$, the knowledge distillation loss $\mathcal{L}_{kd}$, and the KL alignment loss $\mathcal{L}_{kl}$ to update the class-specific model. This enables it to learn discriminative features for new classes while retaining knowledge of previously seen classes. The overall loss for the class-specific branch is defined as:
\begin{equation}\mathcal{L}_{\text{cs}} = \mathcal{L}_{\text{scl}} + \mathcal{L}_{\text{kd}} + \mathcal{L}_{\text{kl}}.\end{equation}

This joint optimization strategy allows the model to not only acquire discriminative features for the current session but also recall informative features learned from previous sessions, thereby enhancing the expressiveness of the class-specific model.
Finally, the total training loss for the overall framework in the incremental learning process is formulated as:
\begin{equation}\mathcal{L}_{\text{total}} = \mathcal{L}_{\text{cs}} + \lambda \cdot \mathcal{L}_{\text{ca}} + \mu \cdot \mathcal{L}_{\text{mcls}},\end{equation}
where $\lambda$ and $\mu$ are hyperparameters that balance the losses $\mathcal{L}_{\text{cs}}$, $\mathcal{L}_{\text{ca}}$, $\mathcal{L}_{\text{mcls}}$. The training objective is to minimize the total loss $\mathcal{L}_{\text{total}}$.

\subsection{Decoupled Classification} 

\subsubsection{Boundary-Aware Exemplar Prioritization.} 

During each incremental learning stage, we employ a sample replay strategy to enrich feature learning with more diverse data. Specifically, we propose a Boundary-Aware Exemplar Prioritization strategy, which selects a representative subset of historical exemplars to be stored in the memory buffer, while simultaneously updating the feature extractor using both old and new class data. Unlike conventional uniform or center-based sampling strategies (e.g., Herding \cite{rebuffi2017icarl, welling2009herding}), BAEP emphasizes the importance of boundary information. Its objective is to select samples that lie near the decision boundary in the feature space and exhibit high representativeness and density. Compared to class-center samples, these boundary exemplars pose greater challenges to the model. This forces the model to learn more generalizable and robust feature representations and enhances its awareness of inter-class separability.

As the number of new classes continues to grow, the computation and storage demand also increases. We define a fixed memory capacity $M$ for the buffer, which stores exemplars from both old and new classes. The storage quota for each class is dynamically adjusted to $k = M / t$, where $t$ denotes the number of classes encountered so far. In each iteration, one sample is selected until the quota $k$ is reached. The selection criterion for the $k$-th exemplar is given by:
\begin{equation}
p_k = \arg\max_{x \in C} \left\| \varphi_\theta(x) - \mu_c + \frac{1}{k} \sum_{j=1}^{k-1} \left( \varphi_\theta(p_j) - \mu_c \right) \right\|_2^2
,\end{equation}
where $C = \{x_1, x_2, \dots, x_n\}$ represents the candidate pool of new class samples, $\varphi_\theta(\cdot)$ is the current feature extractor, and $\mu_c = \frac{1}{n} \sum_{i=1}^n \varphi_\theta(x_i)$ denotes the mean feature embedding of the current class.

BAEP maximizes the deviation of a candidate sample from the class mean and the previously selected exemplars, encouraging the model to preserve boundary instances with strong discriminative power. This strategy ensures that minority classes are sufficiently represented and helps mitigate class imbalance.

\subsubsection{Balanced Random Forest Classifier.} 

In the final classification stage, we adopt the Balanced Random Forest \cite{chen2004using} as a decoupled classifier to predict the final labels based on the features extracted by the class-specific model. Structurally, the BRF is based on traditional random forests but incorporates a class-balanced sampling mechanism, which effectively mitigates the bias introduced by class imbalance.

Specifically, during each BRF construction iteration, suppose the current training set contains $C$ classes. For the $i$-th tree, its training sample subset $S^{\left ( i \right ) }$ consists of the following two parts:
\begin{equation}S^{(i)} = S_{\text{minority}}^{(i)} \cup S_{\text{majority}}^{(i)},\end{equation}
where $S_{\text{minority}}^{(i)}$ denotes the subset obtained by bootstrapping samples from the current minority classes, while $S_{\text{majority}}^{(i)}$ is randomly sampled from the majority classes. Both subsets contain the same number of samples. We then train a CART \cite{breiman2017classification} decision tree $T_{i}$ using this combined subset. At each tree node, feature selection is performed from a random subset $V_{\text{mtry}} \subseteq V$ to enhance the diversity and generalization capability of the tree structure. After training $N$ trees, the BRF classifier makes predictions on a sample $x$ through majority voting:
\begin{equation}
\hat{y}(x) = \text{mode} \bigl( T_1(\varphi(x)), T_2(\varphi(x)), \dots, T_N(\varphi(x)) \bigr),\end{equation}
where $\varphi(x))$ denotes the feature representation of the sample $x$, $T(\cdot)$ indicates the output of the corresponding tree, and $mode(\cdot)$ represents the majority voting operation.

By leveraging structured class-balanced sampling and ensemble voting, the BRF classifier not only alleviates the effect of inter-class distribution shift on final classification but also improves the model’s sensitivity to minority classes.

\section{Experiment}

This section provides a detailed introduction to the two datasets used to evaluate our proposed method: the widely adopted benchmark Tennessee Eastman Process (TEP) dataset \cite{ricker1996decentralized}, and a real-world industrial fault diagnosis dataset, the Multiphase Flow Facility (MFF) \cite{ruiz2015statistical}. We conduct comprehensive evaluations on both datasets to assess the effectiveness of our approach and to validate the contributions of each component through ablation studies.

\subsection{Datasets}

\textbf{TEP Dataset}: In the domain of fault diagnosis, the TEP dataset focuses on simulating realistic chemical processes and has been widely used. It consists of 40 measured variables and 12 manipulated variables (e.g., pressure, temperature, valve opening, and feed flow rates), all of which are continuously monitored by sensors. The TEP dataset consists of 21 types of faults (e.g., step disturbances, valve stiction, etc.), from which we select nine fault types (faults 1, 2, 4, 6, 7, 8, 12, 14, and 18) and one normal class (type 0) to demonstrate the effectiveness of our method under class-imbalanced and long-tailed distributions.

\textbf{MFF Dataset}: The MFF dataset is collected from a gas-liquid-solid three-phase flow separation test platform at Cranfield University. It is a semi-physical dataset that includes 24 process variables (e.g., pressure and flow rate), sampled at 1Hz, with six standard fault types. These variables come from a real-world multiphase flow system. Compared to TEP, the MFF dataset represents a more complex and dynamic industrial process, making it a more challenging benchmark. In our experiments, we randomly select four fault types (types 1, 2, 3, and 4) and one normal type (type 0).

To simulate limited fault data, we restrict the number of samples for each fault class. Following the setting in Peng \cite{peng2022progressively}, class imbalance arises from the normal class having significantly more samples than the fault classes, and long-tailed scenarios are simulated by having a few fault classes with a small number of samples. \tablename~\ref{tab1} shows the class-specific sample settings in the incremental sessions: in the TEP dataset, the number of samples per fault class is set to either 48 or 20, while in the MFF dataset it is set to either 10 or 5.

\begin{table*}[htbp]
  \centering
  \caption{TEP dataset and MFF dataset settings}
    \begin{tabular}{ccccccc}
    \toprule
    \multirow{2}[4]{*}{Dataset} & \multirow{2}[4]{*}{Mode} & \multicolumn{2}{c}{Training Set} & \multicolumn{2}{c}{Testing Set} & \multirow{2}[4]{*}{Incremental Sessions} \\
\cmidrule{3-6}          &       & Normal Class & Fault Class & Normal Class & Fault Class &  \\
    \midrule
    \multirow{2}[2]{*}{TEP} & Imbalanced & \multirow{2}[2]{*}{500} & 48    & \multirow{2}[2]{*}{800} & \multirow{2}[2]{*}{800} & \multirow{2}[2]{*}{5} \\
          & Long-Tailed &       & 20    &       &       &  \\
    \midrule
    \multirow{2}[2]{*}{MFF} & Imbalanced & \multirow{2}[2]{*}{200} & 10    & \multirow{2}[2]{*}{800} & \multirow{2}[2]{*}{800} & \multirow{2}[2]{*}{5} \\
          & Long-Tailed&       & 5     &       &       &  \\
    \bottomrule
    \end{tabular}%
  \label{tab1}%
\end{table*}%

\subsection{Model Setting}

\textbf{Implementation details}: We adopt ResNet18 as the backbone network for both the class-agnostic and class-specific models. For the TEP dataset, we set the memory buffer size to 100 for the imbalanced scenario and 40 for the long-tailed scenario. For the MFF dataset, these values are set to 10 and 5, respectively. All training is conducted on four A40 GPUs using the Adam optimizer. The batch size is set to 512, the weight decay is $1e-5$, and the temperature parameter $\tau$ is set to 0.07. We set the loss balancing hyperparameters $\lambda $ and $\mu $ to 0.5 and 0.6, respectively. The learning rate is fixed at 0.01, and the training lasts for 500 epochs. For data augmentation in the supervised contrastive learning (SCL) stage, we randomly select a segment of the input sequence, shuffle it, and then reintegrate it into the original sequence. The BRF classifier is implemented using the imbalanced-learn \footnote{\url{https://imbalanced-learn.org/stable/references/generated/imblearn.ensemble.BalancedRandomForestClassifier.html\#balancedrandomforestclassifier}} library.

\textbf{Scenario setup}: In our experiments, we focus on multi-stage setups. In these setups, we first train the model on an initial set of tasks $\mathcal{T}_1$, and then sequentially add new classes in subsequent stages $\mathcal{T}_n$. At each stage $\mathcal{T}_n$, we evaluate the model on all classes encountered from stage $\mathcal{T}_1$ to $\mathcal{T}_n$. Specifically, we adopt an incremental class setting of 2+2+2+2+2 classes on the TEP dataset and 1+1+1+1+1+1 classes on the MFF dataset.

\textbf{Evaluation Metrics}: To ensure fair comparison with prior work, we report the average results over the last 10 checkpoints to reduce performance variance. Following previous studies \cite{hell2022data, rebuffi2017icarl, li2017learning, song2023learning}, we adopt classification accuracy and mean per-class accuracy on all classes encountered during incremental learning as our evaluation metrics.

\subsection{Baselines}
To rigorously evaluate the proposed model, we conduct comprehensive comparisons with established baselines and state-of-the-art class-incremental learning methods. Classical approaches include: \textbf{LwF.MC} \cite{li2017learning} which mitigates forgetting through knowledge distillation using previous-task outputs as soft targets without old data access; \textbf{Finetuning} \cite{rebuffi2017icarl} as a simple transfer baseline lacking anti-forgetting mechanisms; \textbf{iCaRL} \cite{rebuffi2017icarl} pioneering rehearsal-based incremental learning via exemplar memory and distillation; \textbf{EEIL} \cite{castro2018end} implementing end-to-end learning with cross-distillation loss; and \textbf{BiC} \cite{wu2019large} correcting classifier bias through linear adjustment. Contemporary state-of-the-art methods comprise: \textbf{SAVC} \cite{song2023learning} enhancing base-class separation with pseudo-supervised contrastive learning; \textbf{WaRP-CIFSL} \cite{kim2023warping} consolidating knowledge by warping parameter spaces; \textbf{BiDistFSCIL} \cite{zhao2023few} preventing overfitting via dual-teacher knowledge transfer; and \textbf{SCLIFD} \cite{peng2023sclifd} tailoring contrastive distillation for industrial incremental scenarios.

\begin{table*}[htbp]
  \centering
  \caption{Comparison of experimental results of TEP and MFF datasets (experimental results retain two decimal places).}
  \label{tab2}%
    \begin{tabular}{ccccccccc}
    \toprule
    \multirow{2}[4]{*}{Dataset} & \multirow{2}[4]{*}{Diagnosis Mode} & \multirow{2}[4]{*}{Method} & \multicolumn{6}{c}{Accuracy(\%) in All Incremental Sessions↑} \\
\cmidrule{4-9}          &       &       & 1     & 2     & 3     & 4     & 5     & Average \\
    \midrule
    \multirow{20}[4]{*}{TEP} & \multirow{10}[2]{*}{Imbalanced} & LwF.MC & \textbf{99.43 } & 42.92  & 18.89  & 13.96  & 10.76  & 37.19  \\
          &       & Finetuing & 99.38  & 47.02  & 32.00  & 15.87  & 17.63  & 42.38  \\
          &       & iCaRL & 98.69  & 72.41  & 60.67  & 58.26  & 54.02  & 68.81  \\
          &       & EEIL  & 58.12  & 33.24  & 17.67  & 14.87  & 11.24  & 27.03  \\
          &       & BiC   & 59.55  & 32.57  & 25.35  & 19.63  & 17.76  & 30.97  \\
          &       & SAVC  & 89.22  & 59.87  & 50.54  & 36.50  & 30.85  & 53.40  \\
          &       & WaRP-CIFSL & 98.35  & 55.54  & 52.48  & 42.59  & 37.84  & 57.36  \\
          &       & BiDistFSCIL & 99.20  & 71.93  & 63.35  & 50.96  & 47.98  & 66.68  \\
          &       & SCLIFD & 99.38  & 76.28  & 92.60  & 82.91  & \textbf{81.42 } & 86.52  \\
          &       & \cellcolor{gray!20}Ours  & \cellcolor{gray!20}99.20  & \cellcolor{gray!20}\textbf{97.44 } & \cellcolor{gray!20}\textbf{97.92 } & \cellcolor{gray!20}\textbf{84.47 } & \cellcolor{gray!20}73.00  & \cellcolor{gray!20}\textbf{90.41 } \\
\cmidrule{2-9}          & \multirow{10}[2]{*}{Long-Tailed} & LwF.MC & \textbf{99.32 } & 44.43  & 26.49  & 14.57  & 13.27  & 37.62  \\
          &       & Finetuing & 99.09  & 46.79  & 31.75  & 14.19  & 15.63  & 41.49  \\
          &       & iCaRL & 98.35  & 70.21  & 58.17  & 54.12  & 51.63  & 66.49  \\
          &       & EEIL  & 57.81  & 27.14  & 13.38  & 10.26  & 8.14  & 23.35  \\
          &       & BiC   & 52.66  & 27.93  & 18.67  & 13.65  & 10.46  & 24.67  \\
          &       & SAVC  & 97.24  & 56.35  & 42.11  & 32.21  & 25.84  & 50.75  \\
          &       & WaRP-CIFSL & 97.27  & 51.31  & 44.40  & 39.01  & 36.68  & 53.73  \\
          &       & BiDistFSCIL & 99.09  & 68.69  & 54.94  & 46.01  & 43.01  & 62.35  \\
          &       & SCLIFD & 99.38  & 79.91  & 94.60  & \textbf{81.74 } & \textbf{74.41 } & \textbf{86.01 } \\
          &       & \cellcolor{gray!20}Ours  & \cellcolor{gray!20}98.92  & \cellcolor{gray!20}\textbf{91.34 } & \cellcolor{gray!20}\textbf{95.38 } & \cellcolor{gray!20}75.29  & \cellcolor{gray!20}64.69  & \cellcolor{gray!20}85.13  \\
    \midrule
    \multirow{20}[4]{*}{MFF} & \multirow{10}[2]{*}{Imbalanced} & LwF.MC & 100.00  & 80.68  & 43.91  & 38.30  & 28.85  & 58.35  \\
          &       & Finetuing & 100.00  & 50.00  & 49.19  & 33.33  & 25.00  & 45.67  \\
          &       & iCaRL & 100.00  & 70.21  & 62.50  & 54.12  & 51.63  & 66.49  \\
          &       & EEIL  & 100.00  & 43.34  & 33.87  & 25.93  & 22.19  & 45.07  \\
          &       & BiC   & 100.00  & 48.45  & 34.25  & 27.81  & 19.23  & 45.95  \\
          &       & SAVC  & 100.00  & 99.94  & 70.50  & 51.78  & 38.80  & 72.20  \\
          &       & WaRP-CIFSL & 100.00  & 97.69  & 89.42  & 96.44  & 97.78  & 97.86  \\
          &       & BiDistFSCIL & 100.00  & 98.19  & 98.63  & 96.72  & 96.18  & 97.94  \\
          &       & SCLIFD & 100.00  & 99.66  & 94.77  & 88.87  & 88.17  & 94.29  \\
          &       & \cellcolor{gray!20}Ours  & \cellcolor{gray!20}\textbf{100.00 } & \cellcolor{gray!20}\textbf{100.00 } & \cellcolor{gray!20}\textbf{100.00 } & \cellcolor{gray!20}\textbf{100.00 } & \cellcolor{gray!20}\textbf{97.92 } & \cellcolor{gray!20}\textbf{99.59 } \\
\cmidrule{2-9}          & \multirow{10}[2]{*}{Long-Tailed} & LwF.MC & 100.00  & 58.47  & 40.20  & 52.38  & 42.31  & 58.67  \\
          &       & Finetuing & 100.00  & 50.00  & 33.33  & 25.00  & 20.00  & 45.67  \\
          &       & iCaRL & 100.00  & 74.94  & 82.13  & 84.56  & 89.93  & 86.31  \\
          &       & EEIL  & 100.00  & 40.52  & 31.62  & 24.56  & 21.25  & 45.51  \\
          &       & BiC   & 100.00  & 41.74  & 30.10  & 21.53  & 17.12  & 42.10  \\
          &       & SAVC  & 100.00  & 93.56  & 95.67  & 72.13  & 56.13  & 83.50  \\
          &       & WaRP-CIFSL & 100.00  & 100.00  & 98.88  & 98.88  & 96.60  & 98.87  \\
          &       & BiDistFSCIL & 100.00  & 99.62  & 100.00  & 96.91  & 88.90  & 97.09  \\
          &       & SCLIFD & 100.00  & 88.18  & 78.52  & 77.98  & 79.81  & 84.90  \\
          &       & \cellcolor{gray!20}Ours  & \cellcolor{gray!20}\textbf{100.00 } & \cellcolor{gray!20}\textbf{100.00 } & \cellcolor{gray!20}\textbf{100.00 } & \cellcolor{gray!20}\textbf{100.00 } & \cellcolor{gray!20}\textbf{99.95 } & \cellcolor{gray!20}\textbf{99.99 } \\
    \bottomrule
    \end{tabular}%
\end{table*}%

\subsection{Results}
\begin{figure*}[!htbp]
    \includegraphics[width=0.9\textwidth]{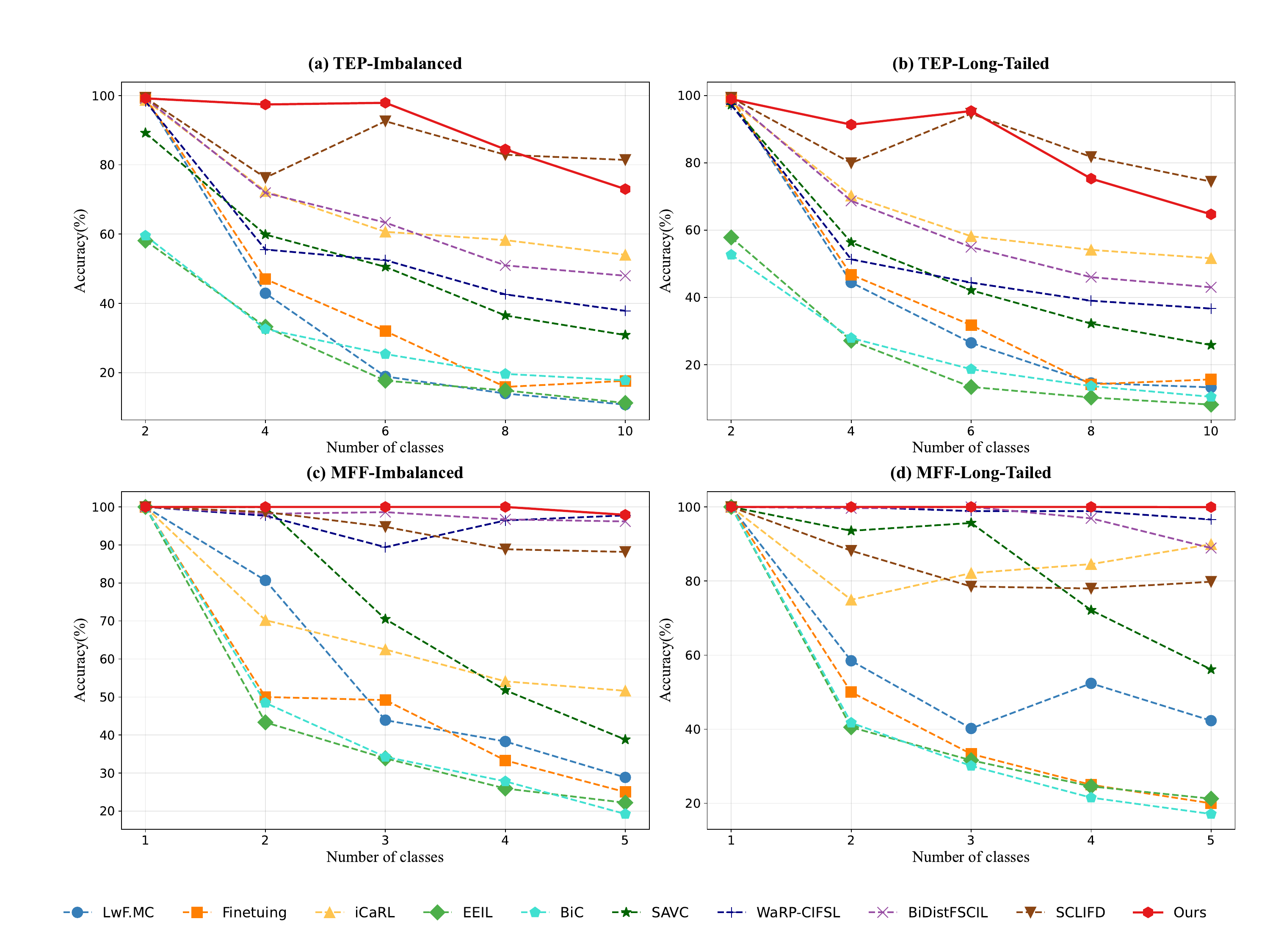}
      \centering
      \caption{Accuracy curves of DGGN and baselines in different incremental scenarios.}
    \label{fig5}
\end{figure*}

This section presents the experimental results of various methods under multi-step incremental learning settings on the TEP and MFF datasets, focusing on the challenging scenarios of class imbalance and long-tailed distributions. As shown in \tablename~\ref{tab2}, we report the classification accuracy at the end of each incremental phase, as well as the overall average accuracy after all incremental sessions. In addressing class imbalance, DGGN consistently achieves higher average classification accuracy than other state-of-the-art methods. For example, compared to the SCLIFD method based on supervised contrastive knowledge distillation, DGGN achieves improvements of 4.51\% and 5.62\% in average class-wise accuracy under class-imbalanced settings on the two datasets, respectively. In the long-tailed scenario on the MFF dataset, DGGN achieves an improvement of 17.77\% in average class-wise accuracy. Furthermore, as the incremental sessions progress, DGGN continues to outperform other methods in terms of accuracy. For instance, in the fifth incremental session on the MFF dataset, DGGN attains an accuracy of 99.95\%, exceeding the second-best method WaRP-CIFSL (96.60\%) by 3.47\%. The superior performance of DGGN can be attributed to the integration of general class-agnostic information, which provides advantages over conventional models that rely on a single feature extractor. Additionally, as shown in \figurename~\ref{fig5}, the accuracy curves of DGGN and baseline methods under various scenarios indicate that DGGN maintains consistently stable performance throughout the entire incremental learning process. This stability demonstrates that our approach effectively mitigates catastrophic forgetting and alleviates the conflict between old and new knowledge in the feature space.

\subsection{Ablation Study}

In our study, the proposed DGGN model significantly enriches feature representation by introducing multi-order interactive aggregation (MOIA) modules and leveraging the universal information from a class-agnostic model. Additionally, we design a multi-semantic cross-attention (MSCA) mechanism to more efficiently integrate features from different semantic domains and to enable knowledge transfer from the class-agnostic model to the class-specific model, thereby enhancing the model’s generalization and discriminative abilities.

To evaluate the independent contributions of 
\noindent these components to class-incremental fault diagnosis, we conduct comprehensive ablation studies under both imbalanced and long-tailed settings on the TEP and MFF datasets. As shown in \tablename~\ref{tab3}, the symbols $\uparrow (\downarrow)$ indicate an improvement (decline) in performance relative to DGGN. The results demonstrate that, when the MOIA module is removed, the average accuracy on the TEP long-tailed task drops from 85.13\% to 83.74\%, and on the MFF long-tailed task from 99.99\% to 97.85\%. This indicates that the multi-order interactive aggregation module plays a critical role in modeling inter-feature relationships in small-sample scenarios. MOIA enables the class-specific model to extract more discriminative features from limited incremental data, thus improving overall diagnostic performance.

Moreover, removing the parallel class-agnostic model causes a substantial decline in accuracy at all incremental stages—most notably, in the TEP imbalanced setting, accuracy drops from 90.41\% to 71.06\%. This result underscores the positive impact of universal semantic representations in guiding the learning of class-specific models. The presence of the class-agnostic model allows our incremental learning framework to achieve better knowledge sharing and generalization between new and old classes, alleviating feature conflicts.

To further investigate the importance of dual-domain feature interaction, we remove the multi-semantic cross-attention module. The results show that the absence of the MSCA module restricts information fusion across semantic domains, with accuracy in the fifth incremental session of the TEP long-tailed task dropping from 64.69\% to 54.13\%. Additionally, when knowledge is not transferred from the class-agnostic model to the class-specific model (i.e., the cross-domain alignment loss is omitted), the accuracy in the MFF imbalanced task decreases by up to 8.24\%.

In summary, the ablation experiments fully validate the effectiveness of each component within our unified framework. These modules work collaboratively to significantly enhance the expressive capacity and stability of the model, which effectively alleviates catastrophic forgetting and feature conflict. As a result, our method demonstrates superior performance and robustness in complex class-incremental fault diagnosis scenarios.

\begin{table*}[htbp]
  \centering
  \caption{Ablation study results on the TEP and MFF datasets. ``MOIA,'' ``CA model,'' ``MSCA,'' and ``w/o Knowledge Transferring'' respectively indicate the model’s multi-order interactive aggregation module, class-agnostic branch, multi-semantic cross-attention module, and the absence of knowledge transfer back to the class-specific model.}
  \label{tab3}%
    \begin{tabular}{ccccccccc}
    \toprule
    \multirow{2}[4]{*}{Dataset} & \multirow{2}[4]{*}{Diagnosis Mode} & \multirow{2}[4]{*}{Method} & \multicolumn{6}{c}{Accuracy(\%) in All Incremental Sessions↑} \\
\cmidrule{4-9}          &       &       & 1     & 2     & 3     & 4     & 5     & Average \\
    \midrule
    \multirow{10}[4]{*}{TEP} & \multirow{5}[2]{*}{Imbalanced} & \cellcolor{gray!20}$\textbf{DGGN}$ & \cellcolor{gray!20}99.20  & \cellcolor{gray!20}97.44  & \cellcolor{gray!20}97.92  & \cellcolor{gray!20}84.47  & \cellcolor{gray!20}73.00  & \cellcolor{gray!20}90.41  \\
          &       & w/o MOIA & 98.98↓ & 97.50↑ & 98.63↑ & 82.01↓ & 68.09↓ & 89.04↓ \\
          &       & w/o CA model & 98.12↓ & 74.61↓ & 72.76↓ & 59.94↓ & 49.84↓ & 71.06↓ \\
          &       & w/o MSCA & 99.15↓ & 98.63↑ & 89.94↓ & 73.00↓ & 57.57↓ & 83.66↓ \\
          &       & w/o Knowledge Transferring & 98.98↓ & 97.89↑ & 97.76↓ & 84.27↓ & 70.60↓ & 89.90↓ \\
\cmidrule{2-9}
          & \multirow{5}[2]{*}{Long-Tailed} & \cellcolor{gray!20}$\textbf{DGGN}$ & \cellcolor{gray!20}98.92  & \cellcolor{gray!20}91.34  & \cellcolor{gray!20}95.38  & \cellcolor{gray!20}75.29  & \cellcolor{gray!20}64.69  & \cellcolor{gray!20}85.13  \\
          &       & w/o MOIA & 99.09↑ & 93.48↑ & 82.46↓ & 72.71↓ & 60.96↓ & 83.74↓ \\
          &       & w/o CA model & 98.98↑ & 72.62↓ & 68.37↓ & 63.20↓ & 51.99↓ & 71.03↓ \\
          &       & w/o MSCA & 98.86↓ & 73.36↓ & 79.92↓ & 62.91↓ & 54.13↓ & 73.84↓ \\
          &       & w/o Knowledge Transferring & 99.03↑ & 83.21↓ & 83.41↓ & 64.95↓ & 54.04↓ & 76.93↓ \\
    \midrule
    \multirow{10}[4]{*}{MFF} & \multirow{5}[2]{*}{Imbalanced} & \cellcolor{gray!20}$\textbf{DGGN}$ & \cellcolor{gray!20}100.00  & \cellcolor{gray!20}100.00  & \cellcolor{gray!20}100.00  & \cellcolor{gray!20}100.00  & \cellcolor{gray!20}97.92  & \cellcolor{gray!20}99.59  \\
          &       & w/o MOIA & 100.00  & 100.00  & 100.00  & 100.00  & 88.75↓ & 97.75↓ \\
          &       & w/o CA model & 100.00  & 100.00  & 100.00  & 77.56↓ & 81.33↓ & 91.78↓ \\
          &       & w/o MSCA & 100.00  & 100.00  & 100.00  & 100.00  & 84.47↓ & 96.90↓ \\
          &       & w/o Knowledge Transferring & 100.00  & 96.81↓ & 100.00  & 100.00  & 89.85↓ & 97.33↓ \\
\cmidrule{2-9}
          & \multirow{5}[2]{*}{Long-Tailed} & \cellcolor{gray!20} $\textbf{DGGN}$ & \cellcolor{gray!20}100.00  & \cellcolor{gray!20}100.00  & \cellcolor{gray!20}100.00  & \cellcolor{gray!20}100.00  & \cellcolor{gray!20}99.95  & \cellcolor{gray!20}99.99  \\
          &       & w/o MOIA & 100.00  & 99.38↓ & 95.04↓ & 96.75↓ & 98.08↓ & 97.85↓ \\
          &       & w/o CA model & 100.00  & 98.38↓ & 98.96↓ & 77.03↓ & 62.65↓ & 87.40↓ \\
          &       & w/o MSCA & 100.00  & 100.00  & 83.04↓ & 97.12↓ & 99.08↓ & 95.85↓ \\
          &       & w/o Knowledge Transferring & 100.00  & 97.44↓ & 97.08↓ & 96.75↓ & 87.33↓ & 95.72↓ \\
    \bottomrule
    \end{tabular}%
\end{table*}%

\begin{table*}[htbp]
  \centering
  \caption{Impact of Multi-Scale Receptive Field Modeling Strategies on Diagnostic Performance and Model Overhead, Where $M1$ denotes $\text{DW}_{7 \times 7}$, $M2$ denotes $\text{DW}_{5 \times 5, d=1} + \text{DW}_{7 \times 7, d=3}$, and $M3$ denotes $\text{DW}_{5 \times 5, d=1} + \text{DW}_{5 \times 5, d=2} + \text{DW}_{7 \times 7, d=3}$. }
  \label{tab4}%
    \begin{tabular}{cccccccccc}
    \toprule
    \multirow{2}[4]{*}{Dataset} & \multirow{2}[4]{*}{Diagnosis Mode} & \multirow{2}[4]{*}{Modules} & \multicolumn{6}{c}{Accuracy(\%) in All Incremental Sessions↑} & \multirow{2}[4]{*}{Params(M)} \\
\cmidrule{4-9}          &       &       & 1     & 2     & 3     & 4     & 5     & Average &  \\
    \midrule
    \multirow{8}[4]{*}{TEP} & \multirow{4}[2]{*}{Imbalanced} & M1 & 99.09  & 98.27  & \textbf{98.12} & 83.86  & 69.85  & 89.84  & 8.50  \\
          &       & M2  & 99.20  & 97.53  & 97.20  & 80.50  & 70.20  & 88.93  & 8.52  \\
          &       & M3 & 98.81  & \textbf{98.36} & 96.51  & 79.79  & 70.40  & 88.77  & 8.54  \\
          &       & \cellcolor{gray!20} $\textbf{DGGN}$ & \cellcolor{gray!20}\textbf{99.20} & \cellcolor{gray!20}97.44  & \cellcolor{gray!20}97.92  & \cellcolor{gray!20}\textbf{84.47} & \cellcolor{gray!20}\textbf{73.00} & \cellcolor{gray!20}\textbf{90.41} & \cellcolor{gray!20}9.91  \\
\cmidrule{2-10}          & \multirow{4}[2]{*}{Long-Tailed} & M1 & 98.86  & 90.39  & 91.13  & 65.59  & 57.92  & 80.78  & 8.50  \\
          &       & M2 & 98.92  & 91.58  & 92.02  & 69.36  & 61.61  & 82.70  & 8.52  \\
          &       & M3 & 98.92  & \textbf{92.11} & 91.29  & 64.09  & 59.41  & 81.16  & 8.54  \\
          &       & \cellcolor{gray!20} $\textbf{DGGN}$ & \cellcolor{gray!20}\textbf{98.92} & \cellcolor{gray!20}91.34  & \cellcolor{gray!20}\textbf{95.38} & \cellcolor{gray!20}\textbf{75.29} & \cellcolor{gray!20}\textbf{64.69} & \cellcolor{gray!20}\textbf{85.13} & \cellcolor{gray!20}9.91  \\
    \bottomrule
    \end{tabular}%
\end{table*}%
            
\subsection{Model Analysis}

In this section,we provide a more in-depth analysis of the importance of DGGN, with all results obtained on the TEP dataset.

\subsubsection{Analysis of the MOIA Module's Role in Modeling Discriminative Features.}
In this section, we conduct an in-depth analysis of the role of the MOIA module in discriminative feature modeling within class-specific models. We perform experiments on the TEP dataset under both class-imbalanced and long-tailed settings, constructing separable convolutions (DWConv) of varying depths. Unlike our proposed parallel multi-branch modeling of multi-order contexts, the comparative structures adopt three typical serial stacking strategies: a single receptive field structure $\left ( \text{DW}_{7 \times 7} \right ) $, a dual-branch serial structure $\left ( \text{DW}_{5 \times 5, d=1} + \text{DW}_{7 \times 7, d=3} \right ) $, and a triple-branch serial structure $\left ( \text{DW}_{5 \times 5, d=1} + \text{DW}_{5 \times 5, d=2} + \text{DW}_{7 \times 7, d=3} \right ) $. 
As shown in \tablename~\ref{tab4}, the MOIA module consistently achieves optimal performance under both experimental settings. In the imbalanced setting, it yields an average accuracy of 90.41\%, which is significantly higher than that of other serial DWConv layers. This advantage is mainly attributed to the superiority of parallel multi-order modeling over serial stacking, as it more effectively captures both local and global contextual dependencies. Although the parallel structure involves slightly more parameters than the serial structure, our model achieves higher classification accuracy in subsequent incremental sessions.

In summary, MOIA enhances the capacity of class-specific models to model complex fault scenarios by constructing parallel low-, mid-, and high-level semantic branches. This results in greater robustness and discriminative power, especially in scenarios with limited samples or semantic ambiguity.

\subsubsection{Comparison with other sample replay strategies.}

\begin{figure*}[!htbp]
    \includegraphics[width=0.9\textwidth]{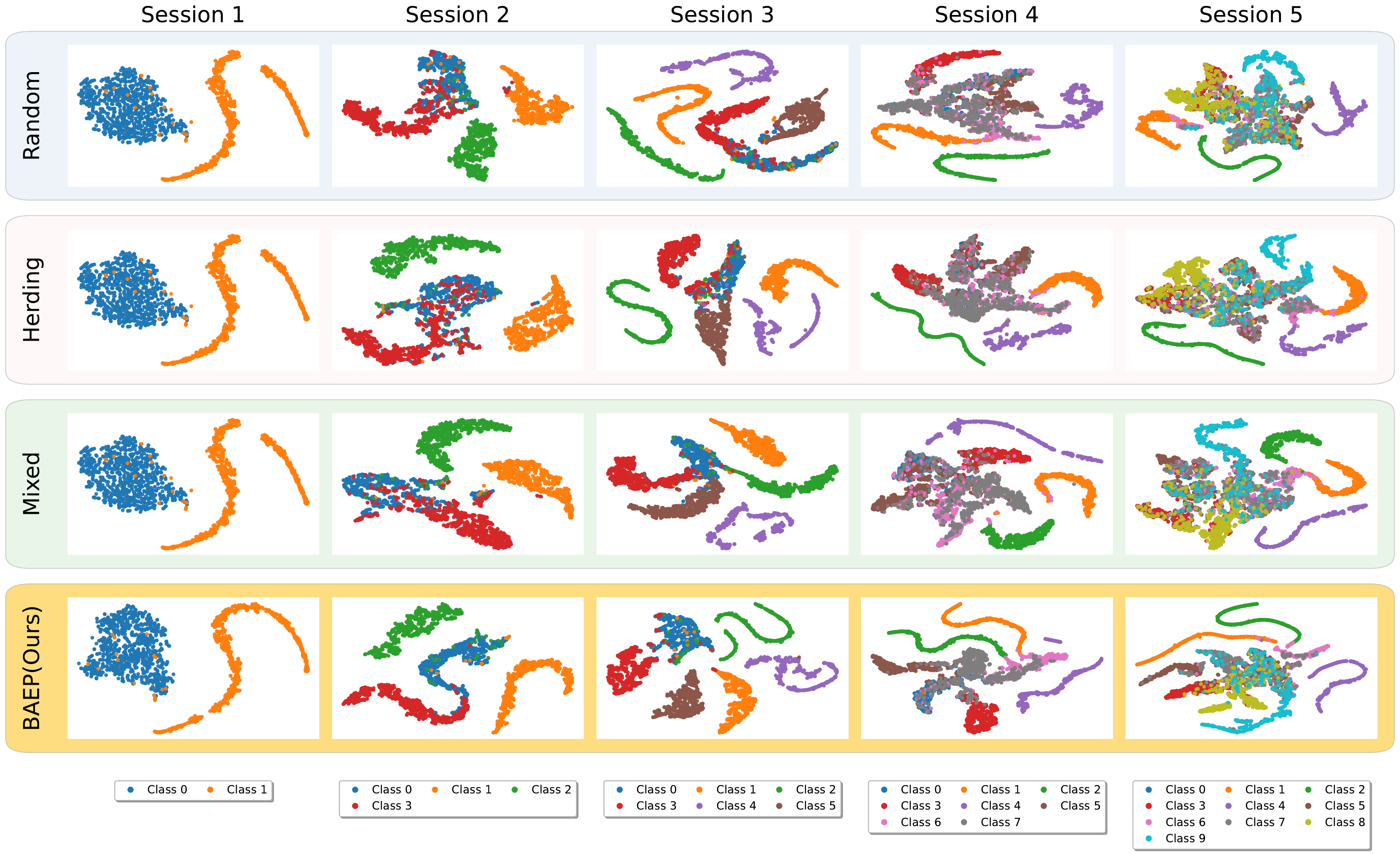}
      \centering
      \caption{t-SNE visualization of class-wise feature representations under different sample replay strategies in class-incremental fault diagnosis tasks.}
    \label{fig6}
\end{figure*}

To further validate the superiority of our sample replay strategy, we compare BAEP with random selection, Herding \cite{welling2009herding}, and a mixed strategy where half of the samples are selected by Herding and the other half by BAEP. We visualize the test feature vectors and their separability at each incremental learning stage using t-SNE, as shown in \figurename~\ref{fig6}. The results demonstrate that the BAEP strategy consistently achieves superior feature separation at every incremental step. Compared with other replay strategies, BAEP plays a positive role in sample selection. For example, in Session 3 and Session 4, BAEP maintains clear decision boundaries for existing classes and provides distinct feature spaces for newly added classes, effectively alleviating the issues of old-class forgetting and new-class interference. In contrast, other strategies show significant class overlap, making it difficult to maintain clear class boundaries. The mixed strategy balances the compactness of Herding and the discriminability of BAEP to some extent, but still falls short of BAEP’s overall performance in feature selection. These results indicate that BAEP is capable of selecting both representative and discriminative samples, thereby providing higher-quality replay samples for class-incremental fault diagnosis.

\begin{table*}[htbp]
  \centering
  \caption{Comparison of Classification Accuracy Across Different Classifiers under Class-Imbalanced and Long-Tailed Scenarios on the TEP Dataset.}
    \begin{tabular}{ccccccccc}
    \toprule
    \multirow{2}[4]{*}{Dataset} & \multirow{2}[4]{*}{Diagnosis Mode} & \multirow{2}[4]{*}{Modules} & \multicolumn{6}{c}{Accuracy(\%) in All Incremental Sessions↑} \\
\cmidrule{4-9}          &       &       & 1     & 2     & 3     & 4     & 5     & Average \\
    \midrule
    \multirow{8}[4]{*}{TEP} & \multirow{4}[2]{*}{Imbalanced} & KNN   & 99.15  & 97.35  & 98.21  & \textbf{85.06 } & 71.56  & 90.26  \\
          &       & SVM   & 99.09  & \textbf{98.48 } & \textbf{98.55 } & 78.28  & 66.10  & 88.10  \\
          &       & FCC   & 78.75  & 94.58  & 91.14  & 71.74  & 65.97  & 80.64  \\
          &       & \cellcolor{gray!20} $\textbf{DGGN}$ & \cellcolor{gray!20}\textbf{99.20 } & \cellcolor{gray!20}97.44  & \cellcolor{gray!20}97.92  & \cellcolor{gray!20}84.47  & \cellcolor{gray!20}\textbf{73.00 } & \cellcolor{gray!20}\textbf{90.41 } \\
\cmidrule{2-9}          & \multirow{4}[2]{*}{Long-Tailed} & KNN   & 98.86  & 88.87  & 90.65  & 70.06  & 61.54  & 82.00  \\
          &       & SVM   & 98.92  & 79.38  & 85.93  & 65.95  & 61.95  & 78.42  \\
          &       & FCC   & 62.33  & 89.64  & 86.79  & 68.83  & 54.74  & 72.47  \\
          &       & \cellcolor{gray!20} $\textbf{DGGN}$ & \cellcolor{gray!20}\textbf{98.92 } & \cellcolor{gray!20}\textbf{91.34 } & \cellcolor{gray!20}\textbf{95.38 } & \cellcolor{gray!20}\textbf{75.29 } & \cellcolor{gray!20}\textbf{64.69 } & \cellcolor{gray!20}\textbf{85.13 } \\
    \bottomrule
    \end{tabular}%
  \label{tab5}%
\end{table*}%

\subsubsection{Applying our approach to other decoupled classifiers.}
To evaluate the performance gain of the BRF classifier, we compare it against several common baseline classifiers, including a fully connected classifier (FCC), K-NN, and SVM. For FCC, a fully connected layer with a Softmax function is attached to the frozen backbone, and the model is trained using cross-entropy loss and the Adam optimizer. All other experimental settings and hyperparameters are kept consistent with those used for BRF. Additionally, the number of neighbors $K$ in K-NN is dynamically set to half the average sample number per class at each session, while the SVM adopts a linear kernel version for computational efficiency in the class-incremental learning setting.
As shown in \tablename~\ref{tab5}, under both settings of the TEP dataset, the BRF classifier consistently outperforms the baselines. In the class-imbalanced scenario, particularly at the fifth incremental session, BRF maintains stronger discriminative power, achieving a classification accuracy of 73\%, whereas KNN and SVM drop to 71.56\% and 66.10\%, respectively. In the more challenging long-tailed scenario, the average accuracy of BRF surpasses those of FCC, KNN, and SVM by 12.66\%, 3.13\%, and 6.71\%, respectively. These results demonstrate the effectiveness of the BRF classifier in maintaining inter-class discriminability and intra-class consistency, thus achieving more accurate classification in incremental learning tasks.

\subsubsection{Dual domain feature similarity analysis.}
\begin{figure}[!htbp]
    \includegraphics[width=\columnwidth]{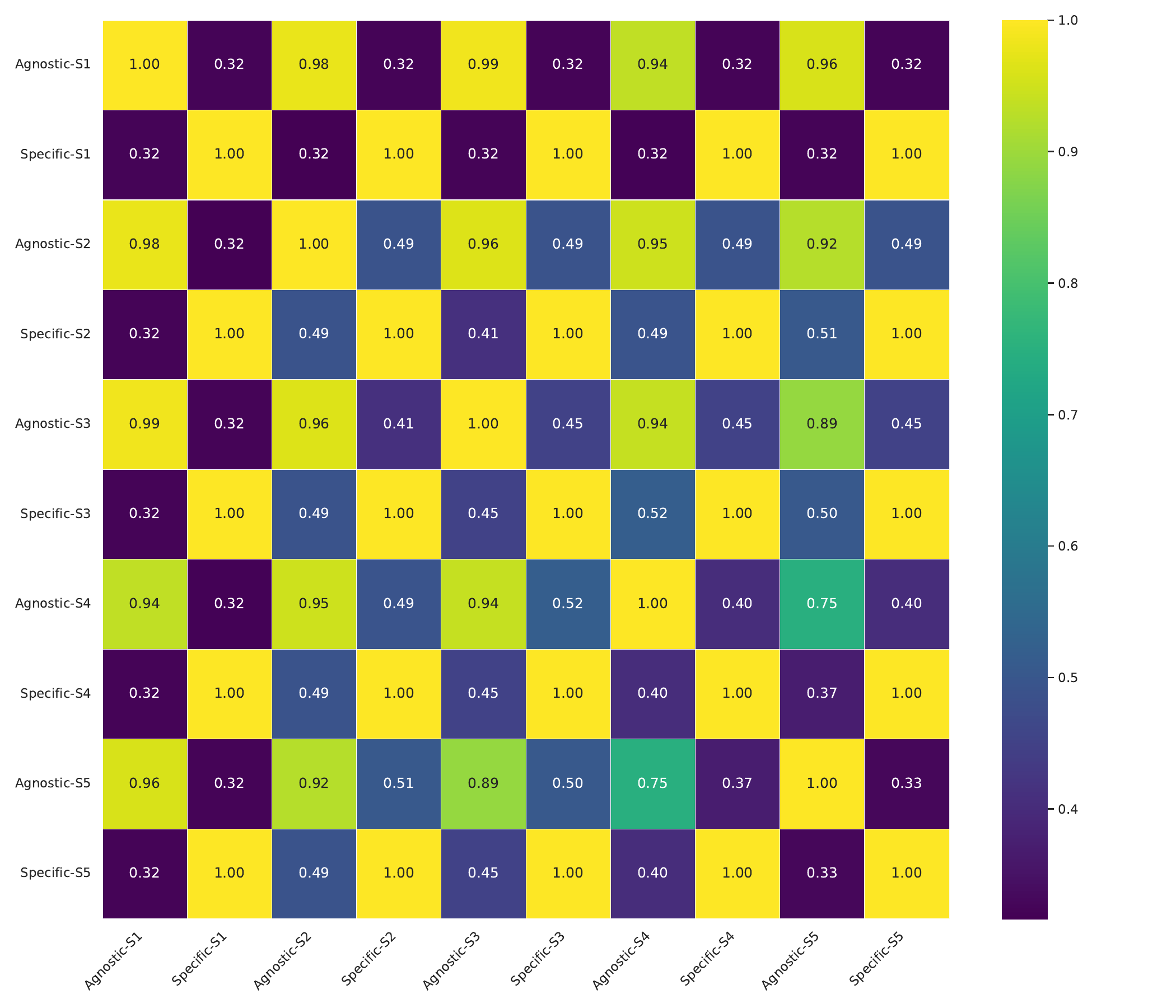}
      \centering
      \caption{CKA similarity heatmap between class-agnostic and class-specific branch features across different incremental sessions.}
    \label{fig7}
\end{figure}

To demonstrate the effectiveness of the learned generic and class-specific feature representations, we adopt centered kernel alignment (CKA)\cite{kornblith2019similarity} as a similarity metric to quantitatively analyze the pairwise feature similarity between the class-agnostic (CA) and class-specific (CS) branches across different incremental learning sessions (S1–S5). We extract CA and CS features from the test set of each session and compute their pairwise CKA similarity, with the results visualized as a heatmap in \figurename~\ref{fig7}. 
As shown in the figure, the CA branch exhibits extremely high feature stability, with inter-session similarities consistently very high (for example, Agnostic-S1 vs. Agnostic-S2 is 0.98, Agnostic-S1 vs. Agnostic-S5 is 0.96, and generally above 0.92). This indicates that the CA branch continually
\begin{table*}[htbp]
  \centering
  \caption{Classification Accuracy for Different Fault Types under the Class-Imbalanced Setting of the TEP Dataset.}
    \begin{tabular}{ccccccccc}
    \toprule
    \multirow{2}[4]{*}{Dataset} & \multirow{2}[4]{*}{Diagnosis Mode} & \multirow{2}[4]{*}{Different Fault Types} & \multicolumn{6}{c}{Accuracy(\%) in All Incremental Sessions↑} \\
\cmidrule{4-9}          &       &       & 1     & 2     & 3     & 4     & 5     & Average \\
    \midrule
    \multirow{4}[2]{*}{TEP} & \multirow{4}[2]{*}{Imbalanced} & 1, 2, 7, 9, 11, 14, 15, 19, 20 & 98.92  & 98.07  & 98.63  & 83.69  & 71.46  & 90.15  \\
          &       & 4, 5, 6, 8, 10, 13, 17, 18, 21 & 99.20  & 98.10  & 98.57  & 81.43  & 71.05  & 89.67  \\
          &       & 1, 3, 5, 7, 12, 14, 16, 19, 20 & 98.81  & 98.63  & 99.09  & 85.26  & 72.72  & 90.90  \\
          &       & 2, 6, 8, 9, 10, 11, 15, 17, 21 & 98.98  & 98.01  & 98.91  & 84.34  & 72.00  & 90.45  \\
    \bottomrule
    \end{tabular}%
  \label{tab6}%
\end{table*}%

\noindent learns and preserves generic, cross-class shared semantics, providing a stable foundation for the model and remaining unaffected by the introduction of new fault classes. In contrast, the CS branch demonstrates clear dynamic evolution in its feature representations, with much lower inter-session similarities (for example, Specific-S2 vs. Specific-S3 is 0.41, Specific-S4 vs. Specific-S5 is 0.37). This suggests that the CS branch constantly adjusts its features as new classes are introduced, to capture the unique discriminative information of the current task. Most importantly, the cross-branch similarity analysis reveals a high degree of disentanglement between the two branches. For example, the similarity between Agnostic-S1 and Specific-S1 within the same session is only 0.32, and that between Agnostic-S1 and Specific-S5 across sessions is around 0.32 as well. These extremely low CKA values strongly demonstrate that the CA and CS branches learn different feature representations, effectively separating generic semantics from class-specific details.

\subsubsection{Results on different fault classes.}

To verify the applicability of our method to other fault types, we further conduct experiments on nine additional fault categories from the 21 fault types in the TEP dataset. These fault categories are divided into four experimental groups, with each group containing a different combination of fault types. The experimental setup remains consistent with previous settings, and we evaluate the classification performance at each incremental stage throughout the entire incremental learning process. As shown in \tablename~\ref{tab6}, under the class-imbalanced setting of the TEP dataset, the average classification accuracy for all four groups exceeds 89.67\%. These results demonstrate that our method maintains high classification accuracy even on fault types beyond the original dataset, further validating the robustness and adaptability of our approach in class-incremental fault diagnosis tasks.

\section{Conclusion}
In this work, we addressed the core challenges in Few-Shot Class-Incremental Fault Diagnosis, namely catastrophic forgetting and overfitting on scarce data. We proposed the Dual-Granularity Guidance Network, a novel framework that constructs parallel class-agnostic and class-specific feature streams. The class-agnostic model captures stable, cross-class knowledge, while the class-specific model focuses on learning discriminative features via multi-order interactive aggregation modules. A multi-semantic cross-attention module aligns these two streams, allowing the general knowledge to guide representation learning and alleviate feature conflicts. Additionally, we employed a Boundary-Aware Exemplar Prioritization strategy and a Balanced Random Forest classifier to further mitigate catastrophic forgetting and bias from data imbalance.

The effectiveness of our approach was validated through extensive experiments on the TEP and MFF datasets, where DGGN consistently demonstrated superior performance and stability over state-of-the-art methods. While the results are strong, future work will focus on three key areas: optimizing the model architecture for greater efficiency, developing dynamic memory management methods, and extending the framework to more complex cross-device or cross-domain scenarios.

\bibliographystyle{IEEEtran}

\bibliography{mybib}

\begin{thebibliography}{10}
\providecommand{\url}[1]{#1}
\csname url@samestyle\endcsname
\providecommand{\newblock}{\relax}
\providecommand{\bibinfo}[2]{#2}
\providecommand{\BIBentrySTDinterwordspacing}{\spaceskip=0pt\relax}
\providecommand{\BIBentryALTinterwordstretchfactor}{4}
\providecommand{\BIBentryALTinterwordspacing}{\spaceskip=\fontdimen2\font plus
\BIBentryALTinterwordstretchfactor\fontdimen3\font minus \fontdimen4\font\relax}
\providecommand{\BIBforeignlanguage}[2]{{%
\expandafter\ifx\csname l@#1\endcsname\relax
\typeout{** WARNING: IEEEtran.bst: No hyphenation pattern has been}%
\typeout{** loaded for the language `#1'. Using the pattern for}%
\typeout{** the default language instead.}%
\else
\language=\csname l@#1\endcsname
\fi
#2}}
\providecommand{\BIBdecl}{\relax}
\BIBdecl

\bibitem{an2023gaussian}
Y.~An, K.~Zhang, Y.~Chai, Z.~Zhu, and Q.~Liu, ``Gaussian mixture variational-based transformer domain adaptation fault diagnosis method and its application in bearing fault diagnosis,'' \emph{IEEE Transactions on Industrial Informatics}, vol.~20, no.~1, pp. 615--625, 2023.

\bibitem{gao2022hierarchical}
Y.~Gao, L.~Gao, X.~Li, and S.~Cao, ``A hierarchical training-convolutional neural network for imbalanced fault diagnosis in complex equipment,'' \emph{IEEE transactions on industrial informatics}, vol.~18, no.~11, pp. 8138--8145, 2022.

\bibitem{li2023order}
M.~Li, P.~Peng, H.~Sun, M.~Wang, and H.~Wang, ``An order-invariant and interpretable dilated convolution neural network for chemical process fault detection and diagnosis,'' \emph{IEEE Transactions on Automation Science and Engineering}, 2023.

\bibitem{wang2024hard}
Z.~Wang, K.~Ma, B.~Qin, J.~Zhang, M.~Li, M.~D. Butala, P.~Peng, and H.~Wang, ``Hard sample mining-enabled supervised contrastive feature learning for wind turbine pitch system fault diagnosis,'' \emph{Measurement Science and Technology}, vol.~35, no.~11, p. 116203, 2024.

\bibitem{wang2024generalized}
X.~Wang, H.~Zhang, X.~Qiao, K.~Ma, S.~Tao, P.~Peng, and H.~Wang, ``Generalized out-of-distribution fault diagnosis (goofd) via internal contrastive learning,'' \emph{IEEE Transactions on Industrial Informatics}, 2024.

\bibitem{mccloskey1989catastrophic}
M.~McCloskey and N.~J. Cohen, ``Catastrophic interference in connectionist networks: The sequential learning problem,'' in \emph{Psychology of learning and motivation}.\hskip 1em plus 0.5em minus 0.4em\relax Elsevier, 1989, vol.~24, pp. 109--165.

\bibitem{li2023incrementally}
C.~Li, X.~Lei, Y.~Huang, F.~Nazeer, J.~Long, and Z.~Yang, ``Incrementally contrastive learning of homologous and interclass features for the fault diagnosis of rolling element bearings,'' \emph{IEEE Transactions on Industrial Informatics}, vol.~19, no.~11, pp. 11\,182--11\,191, 2023.

\bibitem{peng2023sclifd}
P.~Peng, H.~Zhang, M.~Li, G.~Peng, H.~Wang, and W.~Shen, ``Sclifd: Supervised contrastive knowledge distillation for incremental fault diagnosis under limited fault data,'' \emph{arXiv preprint arXiv:2302.05929}, 2023.

\bibitem{chen2022lifelong}
B.~Chen, C.~Shen, D.~Wang, L.~Kong, L.~Chen, and Z.~Zhu, ``A lifelong learning method for gearbox diagnosis with incremental fault types,'' \emph{IEEE transactions on instrumentation and measurement}, vol.~71, pp. 1--10, 2022.

\bibitem{bojian2023continual}
C.~Bojian, S.~Changqing, S.~Juanjuan, K.~Lin, T.~Luyang, W.~Dong, and Z.~Zhongkui, ``Continual learning fault diagnosis: A dual-branch adaptive aggregation residual network for fault diagnosis with machine increments,'' \emph{Chinese Journal of Aeronautics}, vol.~36, no.~6, pp. 361--377, 2023.

\bibitem{zheng2022bearing}
J.~Zheng, H.~Xiong, Y.~Zhang, K.~Su, and Z.~Hu, ``Bearing fault diagnosis via incremental learning based on the repeated replay using memory indexing (r-remind) method,'' \emph{Machines}, vol.~10, no.~5, p. 338, 2022.

\bibitem{zhou2024class}
D.-W. Zhou, Q.-W. Wang, Z.-H. Qi, H.-J. Ye, D.-C. Zhan, and Z.~Liu, ``Class-incremental learning: A survey,'' \emph{IEEE Transactions on Pattern Analysis and Machine Intelligence}, 2024.

\bibitem{belouadah2019il2m}
E.~Belouadah and A.~Popescu, ``Il2m: Class incremental learning with dual memory,'' in \emph{Proceedings of the IEEE/CVF international conference on computer vision}, 2019, pp. 583--592.

\bibitem{hou2019learning}
S.~Hou, X.~Pan, C.~C. Loy, Z.~Wang, and D.~Lin, ``Learning a unified classifier incrementally via rebalancing,'' in \emph{Proceedings of the IEEE/CVF conference on computer vision and pattern recognition}, 2019, pp. 831--839.

\bibitem{rebuffi2017icarl}
S.-A. Rebuffi, A.~Kolesnikov, G.~Sperl, and C.~H. Lampert, ``icarl: Incremental classifier and representation learning,'' in \emph{Proceedings of the IEEE conference on Computer Vision and Pattern Recognition}, 2017, pp. 2001--2010.

\bibitem{zhu2021prototype}
F.~Zhu, X.-Y. Zhang, C.~Wang, F.~Yin, and C.-L. Liu, ``Prototype augmentation and self-supervision for incremental learning,'' in \emph{Proceedings of the IEEE/CVF conference on computer vision and pattern recognition}, 2021, pp. 5871--5880.

\bibitem{chen2022multi}
H.~Chen, Y.~Wang, and Q.~Hu, ``Multi-granularity regularized re-balancing for class incremental learning,'' \emph{IEEE Transactions on Knowledge and Data Engineering}, vol.~35, no.~7, pp. 7263--7277, 2022.

\bibitem{liu2018rotate}
X.~Liu, M.~Masana, L.~Herranz, J.~Van~de Weijer, A.~M. Lopez, and A.~D. Bagdanov, ``Rotate your networks: Better weight consolidation and less catastrophic forgetting,'' in \emph{2018 24th international conference on pattern recognition (ICPR)}.\hskip 1em plus 0.5em minus 0.4em\relax IEEE, 2018, pp. 2262--2268.

\bibitem{yang2021learning}
Y.~Yang, Z.-Q. Sun, H.~Zhu, Y.~Fu, Y.~Zhou, H.~Xiong, and J.~Yang, ``Learning adaptive embedding considering incremental class,'' \emph{IEEE Transactions on Knowledge and Data Engineering}, vol.~35, no.~3, pp. 2736--2749, 2021.

\bibitem{yang2021incremental}
K.~Yang, Z.~Yu, C.~P. Chen, W.~Cao, J.~You, and H.-S. Wong, ``Incremental weighted ensemble broad learning system for imbalanced data,'' \emph{IEEE Transactions on Knowledge and Data Engineering}, vol.~34, no.~12, pp. 5809--5824, 2021.

\bibitem{zhu2021class}
F.~Zhu, Z.~Cheng, X.-y. Zhang, and C.-l. Liu, ``Class-incremental learning via dual augmentation,'' \emph{Advances in neural information processing systems}, vol.~34, pp. 14\,306--14\,318, 2021.

\bibitem{chaudhry2020continual}
A.~Chaudhry, N.~Khan, P.~Dokania, and P.~Torr, ``Continual learning in low-rank orthogonal subspaces,'' \emph{Advances in Neural Information Processing Systems}, vol.~33, pp. 9900--9911, 2020.

\bibitem{wen2020batchensemble}
Y.~Wen, D.~Tran, and J.~Ba, ``Batchensemble: an alternative approach to efficient ensemble and lifelong learning,'' \emph{arXiv preprint arXiv:2002.06715}, 2020.

\bibitem{xiang2023tkil}
J.~Xiang and E.~Shlizerman, ``Tkil: Tangent kernel optimization for class balanced incremental learning,'' in \emph{Proceedings of the IEEE/CVF International Conference on Computer Vision}, 2023, pp. 3529--3539.

\bibitem{wang2025diagllm}
J.~Wang, T.~Li, Y.~Yang, S.~Chen, and W.~Zhai, ``Diagllm: multimodal reasoning with large language model for explainable bearing fault diagnosis,'' \emph{Science China Information Sciences}, vol.~68, no.~6, p. 160103, 2025.

\bibitem{xia2017fault}
M.~Xia, T.~Li, L.~Xu, L.~Liu, and C.~W. De~Silva, ``Fault diagnosis for rotating machinery using multiple sensors and convolutional neural networks,'' \emph{IEEE/ASME transactions on mechatronics}, vol.~23, no.~1, pp. 101--110, 2017.

\bibitem{liu2018fault}
H.~Liu, J.~Zhou, Y.~Zheng, W.~Jiang, and Y.~Zhang, ``Fault diagnosis of rolling bearings with recurrent neural network-based autoencoders,'' \emph{ISA transactions}, vol.~77, pp. 167--178, 2018.

\bibitem{mohammad2023one}
A.~Mohammad-Alikhani, B.~Nahid-Mobarakeh, and M.-F. Hsieh, ``One-dimensional lstm-regulated deep residual network for data-driven fault detection in electric machines,'' \emph{IEEE Transactions on Industrial Electronics}, vol.~71, no.~3, pp. 3083--3092, 2023.

\bibitem{kirkpatrick2017overcoming}
J.~Kirkpatrick, R.~Pascanu, N.~Rabinowitz, J.~Veness, G.~Desjardins, A.~A. Rusu, K.~Milan, J.~Quan, T.~Ramalho, A.~Grabska-Barwinska \emph{et~al.}, ``Overcoming catastrophic forgetting in neural networks,'' \emph{Proceedings of the national academy of sciences}, vol. 114, no.~13, pp. 3521--3526, 2017.

\bibitem{aljundi2018memory}
R.~Aljundi, F.~Babiloni, M.~Elhoseiny, M.~Rohrbach, and T.~Tuytelaars, ``Memory aware synapses: Learning what (not) to forget,'' in \emph{Proceedings of the European conference on computer vision (ECCV)}, 2018, pp. 139--154.

\bibitem{li2017learning}
Z.~Li and D.~Hoiem, ``Learning without forgetting,'' \emph{IEEE transactions on pattern analysis and machine intelligence}, vol.~40, no.~12, pp. 2935--2947, 2017.

\bibitem{shin2017continual}
H.~Shin, J.~K. Lee, J.~Kim, and J.~Kim, ``Continual learning with deep generative replay,'' \emph{Advances in neural information processing systems}, vol.~30, 2017.

\bibitem{lu2022unbalanced}
F.~Lu, Q.~Tong, Z.~Feng, and Q.~Wan, ``Unbalanced bearing fault diagnosis under various speeds based on spectrum alignment and deep transfer convolution neural network,'' \emph{IEEE Transactions on Industrial Informatics}, vol.~19, no.~7, pp. 8295--8306, 2022.

\bibitem{zhang2024multiscale}
X.~Zhang, J.~Liu, X.~Zhang, and Y.~Lu, ``Multiscale channel attention-driven graph dynamic fusion learning method for robust fault diagnosis,'' \emph{IEEE Transactions on Industrial Informatics}, vol.~20, no.~9, pp. 11\,002--11\,013, 2024.

\bibitem{khosla2020supervised}
P.~Khosla, P.~Teterwak, C.~Wang, A.~Sarna, Y.~Tian, P.~Isola, A.~Maschinot, C.~Liu, and D.~Krishnan, ``Supervised contrastive learning,'' \emph{Advances in neural information processing systems}, vol.~33, pp. 18\,661--18\,673, 2020.

\bibitem{fini2022self}
E.~Fini, V.~G.~T. Da~Costa, X.~Alameda-Pineda, E.~Ricci, K.~Alahari, and J.~Mairal, ``Self-supervised models are continual learners,'' in \emph{Proceedings of the IEEE/CVF conference on computer vision and pattern recognition}, 2022, pp. 9621--9630.

\bibitem{chen2020simple}
T.~Chen, S.~Kornblith, M.~Norouzi, and G.~Hinton, ``A simple framework for contrastive learning of visual representations,'' in \emph{International conference on machine learning}.\hskip 1em plus 0.5em minus 0.4em\relax PmLR, 2020, pp. 1597--1607.

\bibitem{welling2009herding}
M.~Welling, ``Herding dynamical weights to learn,'' in \emph{Proceedings of the 26th annual international conference on machine learning}, 2009, pp. 1121--1128.

\bibitem{chen2004using}
C.~Chen, A.~Liaw, L.~Breiman \emph{et~al.}, ``Using random forest to learn imbalanced data,'' \emph{University of California, Berkeley}, vol. 110, no. 1-12, p.~24, 2004.

\bibitem{breiman2017classification}
L.~Breiman, J.~Friedman, R.~A. Olshen, and C.~J. Stone, \emph{Classification and regression trees}.\hskip 1em plus 0.5em minus 0.4em\relax Routledge, 2017.

\bibitem{ricker1996decentralized}
N.~L. Ricker, ``Decentralized control of the tennessee eastman challenge process,'' \emph{Journal of process control}, vol.~6, no.~4, pp. 205--221, 1996.

\bibitem{ruiz2015statistical}
C.~Ruiz-C{\'a}rcel, Y.~Cao, D.~Mba, L.~Lao, and R.~Samuel, ``Statistical process monitoring of a multiphase flow facility,'' \emph{Control Engineering Practice}, vol.~42, pp. 74--88, 2015.

\bibitem{peng2022progressively}
P.~Peng, J.~Lu, S.~Tao, K.~Ma, Y.~Zhang, H.~Wang, and H.~Zhang, ``Progressively balanced supervised contrastive representation learning for long-tailed fault diagnosis,'' \emph{IEEE Transactions on Instrumentation and Measurement}, vol.~71, pp. 1--12, 2022.

\bibitem{hell2022data}
M.~Hell, E.~Pestana~de Aguiar, N.~Soares, and L.~Goliatt, ``A data-driven time-series fault prediction framework for dynamically evolving large-scale data streaming systems,'' \emph{International Journal of Fuzzy Systems}, vol.~24, no.~6, pp. 2831--2844, 2022.

\bibitem{song2023learning}
Z.~Song, Y.~Zhao, Y.~Shi, P.~Peng, L.~Yuan, and Y.~Tian, ``Learning with fantasy: Semantic-aware virtual contrastive constraint for few-shot class-incremental learning,'' in \emph{Proceedings of the IEEE/CVF conference on computer vision and pattern recognition}, 2023, pp. 24\,183--24\,192.

\bibitem{castro2018end}
F.~M. Castro, M.~J. Mar{\'\i}n-Jim{\'e}nez, N.~Guil, C.~Schmid, and K.~Alahari, ``End-to-end incremental learning,'' in \emph{Proceedings of the European conference on computer vision (ECCV)}, 2018, pp. 233--248.

\bibitem{wu2019large}
Y.~Wu, Y.~Chen, L.~Wang, Y.~Ye, Z.~Liu, Y.~Guo, and Y.~Fu, ``Large scale incremental learning,'' in \emph{Proceedings of the IEEE/CVF conference on computer vision and pattern recognition}, 2019, pp. 374--382.

\bibitem{kim2023warping}
D.-Y. Kim, D.-J. Han, J.~Seo, and J.~Moon, ``Warping the space: Weight space rotation for class-incremental few-shot learning,'' in \emph{The Eleventh International Conference on Learning Representations}, 2023.

\bibitem{zhao2023few}
L.~Zhao, J.~Lu, Y.~Xu, Z.~Cheng, D.~Guo, Y.~Niu, and X.~Fang, ``Few-shot class-incremental learning via class-aware bilateral distillation,'' in \emph{Proceedings of the IEEE/CVF conference on computer vision and pattern recognition}, 2023, pp. 11\,838--11\,847.

\bibitem{kornblith2019similarity}
S.~Kornblith, M.~Norouzi, H.~Lee, and G.~Hinton, ``Similarity of neural network representations revisited,'' in \emph{International conference on machine learning}.\hskip 1em plus 0.5em minus 0.4em\relax PMLR, 2019, pp. 3519--3529.

\end{thebibliography}

\newpage

\end{document}